\documentclass[12pt, onecolumn]{IEEEtran}

\usepackage{siunitx}
\usepackage{booktabs}
\usepackage{graphicx}
\usepackage{multirow}
\usepackage{soul}
\usepackage{amssymb}
\usepackage{amsmath}
\graphicspath{{figure/}}

\newcommand{\ie}{\textit{i}.\textit{e}., }
\newcommand{\eg}{\textit{e}.\textit{g}., }
\newcommand{\etc}{\textit{etc}}

\newcommand{\etal}{\textit{et al}. }

\usepackage{hyperref}

\begin{document}
	
	\title{Federated Reinforcement Learning: Techniques, Applications, and Open Challenges}
	
	\author{Jiaju Qi, Qihao Zhou, Lei Lei, Kan Zheng}
	
	\maketitle

	\begin{abstract}
		
		
		This paper presents a comprehensive survey of Federated Reinforcement Learning (FRL), an emerging and promising field in Reinforcement Learning (RL). Starting with a tutorial of Federated Learning (FL) and RL, we then focus on the introduction of FRL as a new method with great potential by leveraging the basic idea of FL to improve the performance of RL while preserving data-privacy. According to the distribution characteristics of the agents in the framework, FRL algorithms can be divided into two categories, \ie Horizontal Federated Reinforcement Learning (HFRL) and Vertical Federated Reinforcement Learning (VFRL). We provide the detailed definitions of each category by formulas, investigate the evolution of FRL from a technical perspective, and highlight its advantages over previous RL algorithms. In addition, the existing works on FRL are summarized by application fields, including edge computing, communication, control optimization, and attack detection. Finally, we describe and discuss several key research directions that are crucial to solving the open problems within FRL.
		
		
	\end{abstract}
	
	
	\section{Introduction}
	
	As Machine Learning (ML) develops, it becomes capable of solving increasingly complex problems, such as image recognition, speech recognition, and semantic understanding.  Despite the effectiveness of traditional machine learning algorithms in several areas, the researchers found that scenes involving many parties are still difficult to resolve, especially when privacy is concerned. Federated Learning (FL), in these cases, has attracted increasing interest among ML researchers. Technically, the FL is a decentralized collaborative approach that allows multiple partners to train data respectively and build a shared model while maintaining privacy. With its innovative learning architecture and concepts, FL provides safer experience exchange services and enhances capabilities of ML in distributed scenarios.
	
	In ML, Reinforcement Learning (RL) is one of the branches that focuses on how individuals, \ie agents, interact with their environment and maximize some portion of the cumulative reward. The process allows agents to learn to improve their behavior in a trial and error manner. Through a set of policies, they take actions to explore the environment and expect to be rewarded. Research on RL has been hot in recent years, and it has shown great potential in various applications, including games, robotics, communication, and so on.
	
	However, there are still many problems in the implementation of RL in practical scenarios.  For example, considering that in the case of large action space and state space, the performance of agents is vulnerable to collected samples since it is nearly impossible to explore all sampling spaces. In addition, many RL algorithms have the problem of learning efficiency caused by low sample efficiency. Therefore, through information exchange between agents, learning speed can be greatly accelerated. Although distributed RL and parallel RL algorithms \cite{DBLP:journals/corr/NairSBAFMPSBPLM15, 10.1007/978-3-540-77949-0_5,DBLP:journals/corr/ClementeMC17} can be used to address the above problems, they usually need to collect all the data, parameters, or gradients from each agent in a central server for model training. However, one of the important issues is that some tasks need to prevent agent information leakage and protect agent privacy during the application of RL. Agents' distrust of the central server and the risk of eavesdropping on the transmission of raw data has become a major bottleneck for such RL applications. FL can not only complete information exchange while avoiding privacy disclosure, but also adapt various agents to their different environments. Another problem of RL is how to bridge the simulation-reality gap. Many RL algorithms require pre-training in simulated environments as a prerequisite for application deployment, but one problem is that the simulated environments cannot accurately reflect the environments of the real world. FL can aggregate information from both environments and thus bridge the gap between them.  Finally, in some cases, only partial features can be observed by each agent in RL. However, these features, no matter observations or rewards, are not enough to obtain sufficient information required to make decisions. At this time, FL makes it possible to integrate this information through aggregation.
	
	Thus, the above challenges give rise to the idea of federated reinforcement learning (FRL). As FRL can be considered as an integration of FL and RL under privacy protection, several elements of RL can be presented in FL frameworks to deals with sequential decision-making tasks. For example, these three dimensions of sample, feature and label in FL can be replaced by environment, state and action respectively in FRL. Since FL can be divided into several categories according to the distribution characteristics of data, including Horizontal Federated Learning (HFL) and Vertical Federated Learning (VFL), we can similarly categorize FRL algorithms into Horizontal Federated Reinforcement Learning (HFRL) and Vertical Federated Reinforcement Learning (VFRL).
	
	
	Though a few survey papers on FL \cite{9060868,9415623,khan2021federated} have been published, to the best of our knowledge, there are currently no relevant survey papers focused on FRL. Due to the fact that FRL is a relatively new technique, most researchers may be unfamiliar with it to some extent. We hope to identify achievements from current studies and serve as a stepping stone to further research. In summary, this paper sheds light on the following aspects.
	
	\begin{enumerate}
		\item \emph{Systematic tutorial on FRL methodology}. As a review focusing on FRL, this paper tries to explain the knowledge about FRL to researchers systematically and in detail. The definition and categories of FRL are introduced firstly, including system model, algorithm process, \etc. In order to explain the framework of HFRL and VFRL and the difference between them clearly, two specific cases are introduced, \ie autonomous driving and smart grid. Moreover, we comprehensively introduce the existing research on FRL's algorithm design.
		\item \emph{Comprehensive summary for FRL applications}. This paper collects a large number of references in the field of FRL, and provides a comprehensive and detailed investigation of the FRL applications in various areas, including edge computing, communications, control optimization, attack detection, and some other applications. For each reference, we discuss the authors’ research ideas and methods, and summarize how the researchers combine the FRL algorithm with the specific practical problems.
		\item \emph{Open issues for future research}. This paper identifies several open issues for FRL as a guide for further research. The scope covers communication, privacy and security, join and exit mechanisms design, learning convergence and some other issues. We hope that they can broaden the thinking of interested researchers and provide help for further research on FRL.
	\end{enumerate}
	
	The organization of this paper is as follows. To quickly gain a comprehensive understanding of FRL, the paper starts with FL and RL in Section 2 and Section 3, respectively, and extends the discussion further to FRL in Section 4. The existing applications of FRL are summarized in Section 5. In addition, a few open issues and future research directions for FRL are highlighted in Section 6. Finally, the conclusion is given in Section 7.
	
	
	\section{Federated learning}
	\subsection{Federated learning definition and basics}
	In general, FL is a ML algorithmic framework that allows multiple parties to perform ML under the requirements of privacy protection, data security, and regulations \cite{bookfederatedlearning}. In FL architecture, model construction includes two processes: model training and model inference. It is possible to exchange information about the model between parties during training, but not the data itself, so that data privacy will not be compromised in any way. An individual party or multiple parties can possess and maintain the trained model. In the process of model aggregation, more data instances collected from various parties contribute to updating the model. As the last step, a fair value-distribution mechanism should be used to share the profits obtained by the collaborative model \cite{yang2019federated}. The well-designed mechanism enables the federation sustainability. Aiming to build a joint ML model without sharing local data, FL involves technologies from different research fields such as distributed systems, information communication, ML and cryptography \cite{DBLP:journals/corr/abs-1907-09693}. FL has the following characteristics as a result of these techniques, \ie
	\begin{itemize}
		\item Distribution. There are two or more parties that hope to jointly build a model to tackle similar tasks. Each party holds independent data and would like to use it for model training.
		\item Data protection. The data held by each party does not need to be sent to the other during the training of the model. The learned profits or experiences are conveyed through model parameters that do not involve privacy.
		\item Secure communication. The model is able to be transmitted between parties with the support of an encryption scheme. The original data cannot be inferred even if it is eavesdropped during transmission.
		\item Generality. It is possible to apply FL to different data structures and institutions without regard to domains or algorithms. 
		\item Guaranteed performance. The performance of the resulting model is very close to that of the ideal model established with all data transferred to one centralized party.
		\item Status equality. To ensure the fairness of cooperation, all participating parties are on an equal footing. The shared model can be used by each party to improve its local models when needed.
	\end{itemize}
	A formal definition of FL is presented as follows. Consider that there are $N$ parties $ \left\{ \mathcal{F} _i \right\} _{i=1}^{N} $ interested in establishing and training a cooperative ML model. Each party has their respective datasets  $\mathcal{D} _i$. Traditional ML approaches consist of collecting all data $\left\{ \mathcal{D} _i \right\} _{i=1}^{N}$  together to form a centralized dataset $\mathbb{R} $  at one data server. The expected model $\mathcal{M} _{SUM}$  is trained by using the dataset $\mathbb{R} $. On the other hand, FL is a reform of ML process in which the participants $\mathcal{F} _i$  with data $\mathcal{D} _i$ jointly train a target model $\mathcal{M} _{FED}$  without aggregating their data. Respective data  $\mathcal{D} _i$ is stored on the owner $\mathcal{F} _i$  and not exposed to others. In addition, the performance measure of the federated model $\mathcal{M} _{FED}$ is denoted as $\mathcal{V} _{FED}$, including accuracy, recall, and F1-score, \etc, which should be a good approximation of the performance of the expected model $\mathcal{M} _{SUM}$, \ie $\mathcal{V} _{SUM}$. In order to quantify differences in performance, let $\delta $ be a non-negative real number and define the federated learning model $\mathcal{M} _{FED}$  has $\delta $ performance loss if 
	$$
	|\mathcal{V} _{SUM}-\mathcal{V} _{FED}|<\delta .
	$$

	Specifically,  the FL model hold by each party is basically the same as the ML model, and it also includes a set of parameters $w_{i}$ which is learned based on the respective training dataset $\mathcal{D} _i$ \cite{9084352}. A training sample $j$ typically contains both the input of FL model and the expected output. For example, in the case of image recognition, the input is the pixel of the image, and the expected output is the correct label. The learning process is facilitated by defining a loss function on  parameter vector $w$ for every data sample $j$. The loss function represents the error of the model in relation to the training data. For each dataset $\mathcal{D} _i$ at party $ \mathcal{F} _i$, the loss function on the collection of training samples can be defined as follow \cite{8664630},
	
	$$
	F_i\left( w \right) =\frac{1}{\left| \mathcal{D} _i \right|}\sum_{j\in \mathcal{D} _i}^{}{f_j\left( w \right)},
	$$
	
	where $f_j\left( w \right)$ denotes the loss function of the sample $j$ with the given model parameter vector $w$ and $|\cdot|$ represents the size of the set. In FL, it is important to define the global loss function since multiple parties are jointly training a global statistical model without sharing a dataset. The common global loss function on all the distributed datasets is given by,
	$$
	F_g\left( w \right) = \sum_{i=1}^{N}{\eta_i F_i\left( w \right)},
	$$
	
	where $\eta_i$ indicates the relative impact of each party on the global model. In addition, $\eta_i > 0$ and $\sum_{i=1}^{N} \eta_i = 1$. This term $\eta$ can be flexibly defined to  improve training efficiency. The natural setting is averaging between parties, \ie $\eta = 1/N$. The goal of the learning problem is to find the optimal parameter that minimizes the global loss function $F_g\left( w \right)$. It is presented in formula form,
	$$
	w^* = \mathop{\mathrm{arg\,min}}_w F_g\left( w \right).
	$$
	
	Considering that FL is designed to adapt to various scenarios, the objective function may be appropriate depending on the application. However, a closed-form solution is almost impossible to find with most FL models due to their inherent complexity. A canonical federated averaging algorithm (FedAvg) based on gradient-descent techniques is presented in the study from McMahan \etal \cite{DBLP:journals/corr/McMahanMRA16}, which is widely used in FL systems. In general, the coordinator has the initial FL model and is responsible for aggregation. Distributed participants know the optimizer settings and can upload information that does not affect privacy. The specific architecture of FL will be discussed in the next subsection. Each participant uses their local data to perform one step (or multiple steps) of gradient descent on the current model parameter $\bar{w}\left( t \right)$ according to the following formula,
	$$
	\forall i, w_i\left( t+1 \right) =\bar{w}\left( t \right) -\gamma \nabla F_i\left( \bar{w}_i\left( t \right) \right), 
	$$
	
	where $\gamma$ denotes a fixed learning rate of each gradient descent. After receiving the local parameters from participants, the central coordinator updates the global model using a weighted average, \ie
	$$
	\bar{w}_g\left( t+1 \right) =\sum_{i=1}^N{\frac{n_i}{n}w_i\left( t+1 \right)},
	$$
	
	where $n_i$ indicates the number of training data samples of the $i$-th participant has and $n$ denotes the total number of samples contained in all the datasets. Finally, the coordinator sends the aggregated model weights $\bar{w}_g\left( t+1 \right)$ back to the participants. The aggregation process is performed at a predetermined interval or iteration round. Additionally, FL leverages privacy-preserving techniques to prevent the leakage of gradients or model weights. For example, the existing encryption algorithms are added on top of the original FedAvg to provide secure FL \cite{8241854,8744465}.
	
	\subsection{Architecture of federated learning}
	According to the application characteristics, the architecture of FL can be divided into two types \cite{bookfederatedlearning}, i.e., client-server model and peer-to-peer model.
	
\begin{figure}[htb]
		\centering
		\includegraphics[width=0.6\textwidth]{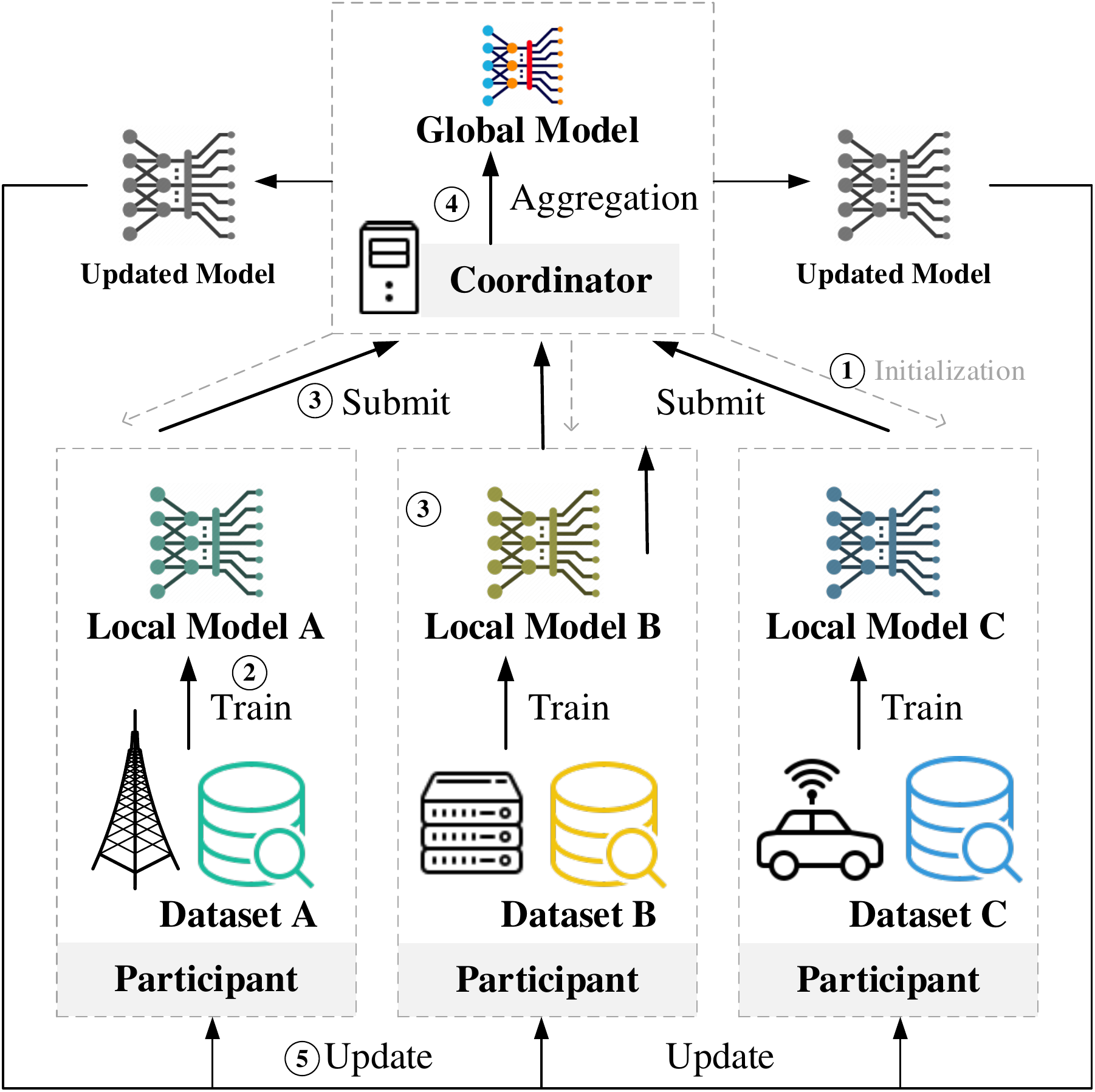}
		\caption{An Example of Federated Learning Architecture: Client-Server Model.}
		\label{fig-clint-server}
	\end{figure}

	As shown in Figure \ref{fig-clint-server}, there are two major components in the client-server model, \ie participants and coordinators. The participants are the data owners and can perform local model training and updates. In different scenarios, the participants are made up of different devices,  the vehicles in the Internet of Vehicles (IoV), or the smart devices in the IoT. In addition, participants usually possess at least two characteristics. Firstly, each participant has a certain level of hardware performance, including computation power, communication and storage. The hardware capabilities ensure that the FL algorithm operates normally. Secondly, participants are independent of one another and located in a wide geographic area. In the client-server model, coordinator can be considered as a central aggregation server, which can initialize a model and aggregate model updates from participants \cite{DBLP:journals/corr/McMahanMRA16}. As participants train both based on local data sets concurrently and share their experience through the coordinator with the model aggregation mechanism, it will greatly enhance the efficiency of the training and enhance the performance of the model. However, since participants won't be able to communicate directly, the coordinator must perform well to train the global model and maintain communication with all participants. Therefore, the model has security challenges such as a single point of failure. If the coordinator fails to complete the model aggregation task, the local model of participant has difficulty meeting target performance. The basic workflow of the client-server model can be summarized in the following five steps. The process continues to repeat the steps from 2 to 5 until the model converges, or until the maximum number of iterations is reached.
	\begin{itemize}
		\item Step 1: In the process of setting up a client-server-based learning system, the coordinator creates an initial model and sends it to each participant. Those participants who join later can access the latest global model. 
		\item Step 2: Each participant trains a local model based on their respective dataset.
		\item Step 3: Updates of model parameters are sent to the central coordinator.
		\item Step 4: The coordinator combines the model updates using specific aggregation algorithms.
		\item Step 5: The combined model is sent back to the corresponding participant.
	\end{itemize}
	
\begin{figure}[htb]
		\centering
		\includegraphics[width=0.6\textwidth]{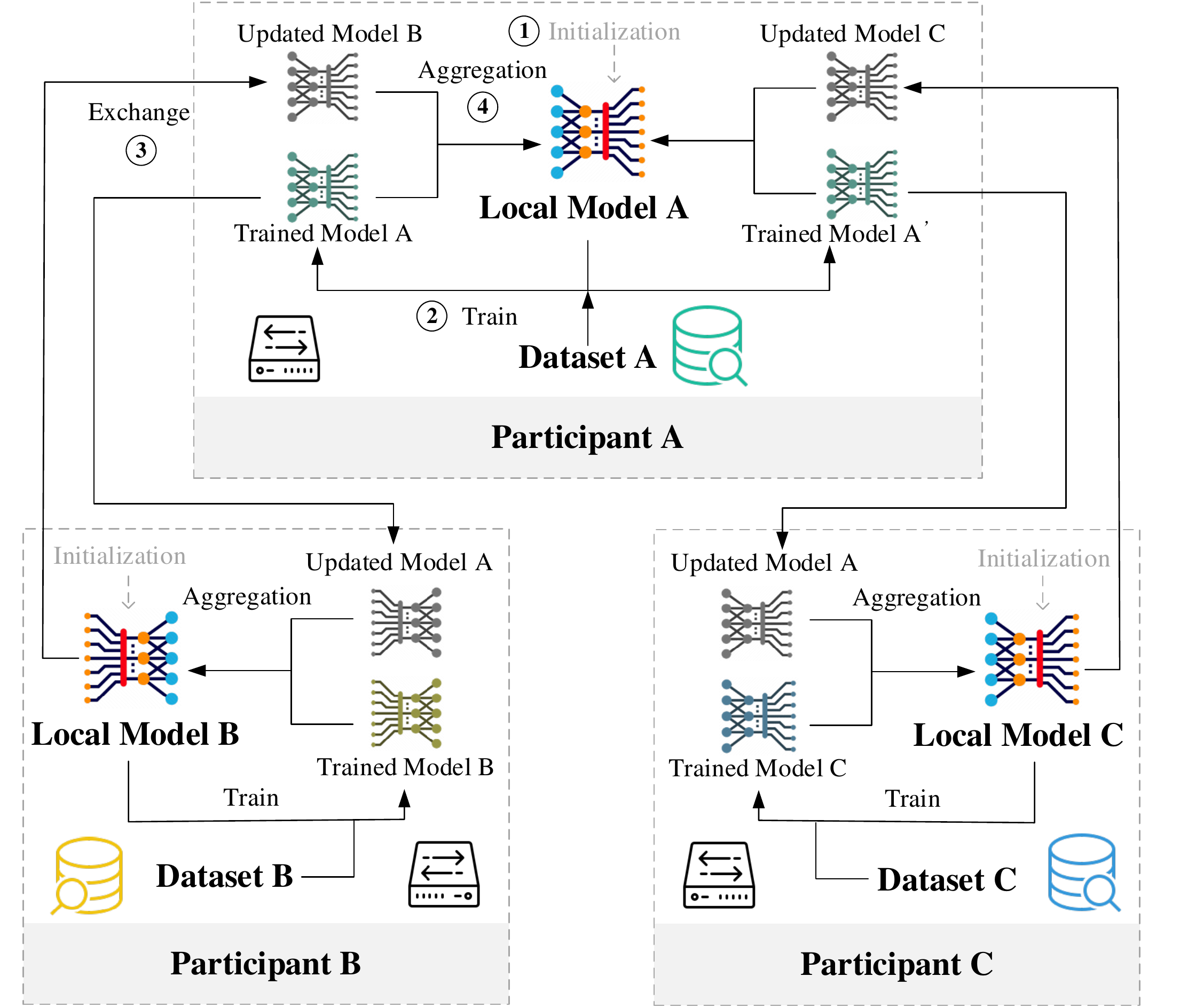}
		\caption{An Example of Federated Learning Architecture: Peer-to-Peer Model.}
		\label{fig-peer-to-peer}
	\end{figure}

	The peer-to-peer based FL architecture does not require a coordinator as illustrated in Figure \ref{fig-peer-to-peer}. Participants can directly communicate with each other without relying on a third party. Therefore, each participant in the architecture is equal and can initiate a model exchange request with anyone else. As there is no central server, participants must agree in advance on the order in which model should be sent and received. Common transfer modes are cyclic transfer and random transfer. The peer-to-peer model is suitable and important for specific scenarios. For example, multiple banks jointly develop an ML-based attack detection model. With FL, there is no need to establish a central authority between banks to manage and store all attack patterns. The attack record is only held at the server of the attacked bank, but the detection experience can be shared with other participants through model parameters. The FL procedure of the peer-to-peer model is simpler than that of the client-server model. 
	\begin{itemize}
		\item Step 1: Each participant initializes their local model depending on its needs. 
		\item Step 2: Train the local model based on the respective dataset.
		\item Step 3: Create a model exchange request to other participants and send local model parameters.
		\item Step 4: Aggregate the model received from other participants into the local model.
	\end{itemize}
	The termination conditions of the process can be designed by participants according to their needs. This architecture further guarantees security since there is no centralized server. However, it requires more communication resources and potentially increased computation for more messages.
	\subsection{Categories of federated learning}
\begin{figure}[htb]
		\centering
		\includegraphics[width=0.57\textwidth]{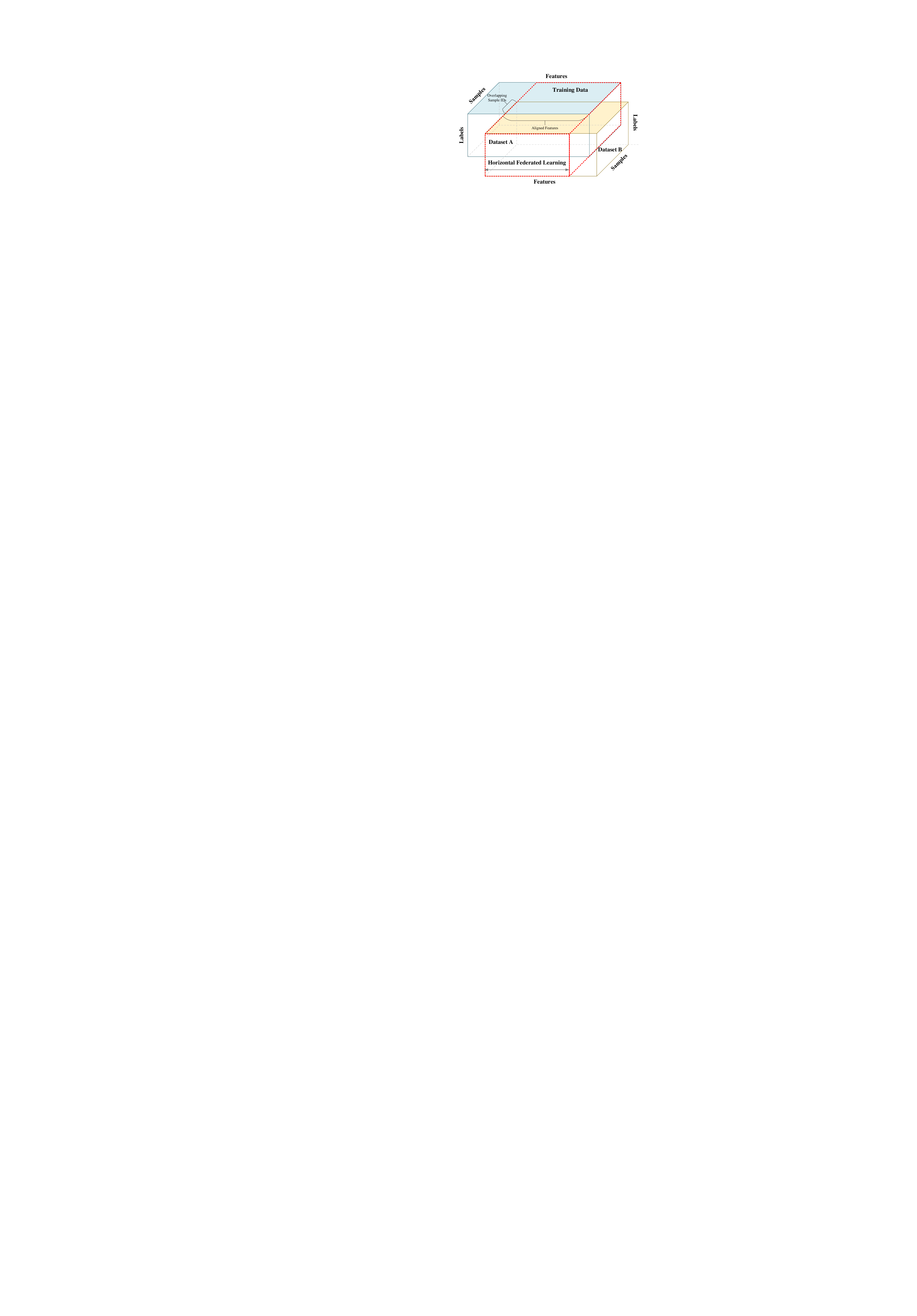}
		\caption{Illustration of horizontal federated learning (HFL).}
		\label{fig-horizontal}
	\end{figure}

\begin{figure}[htb]
		\centering
		\includegraphics[width=0.6\textwidth]{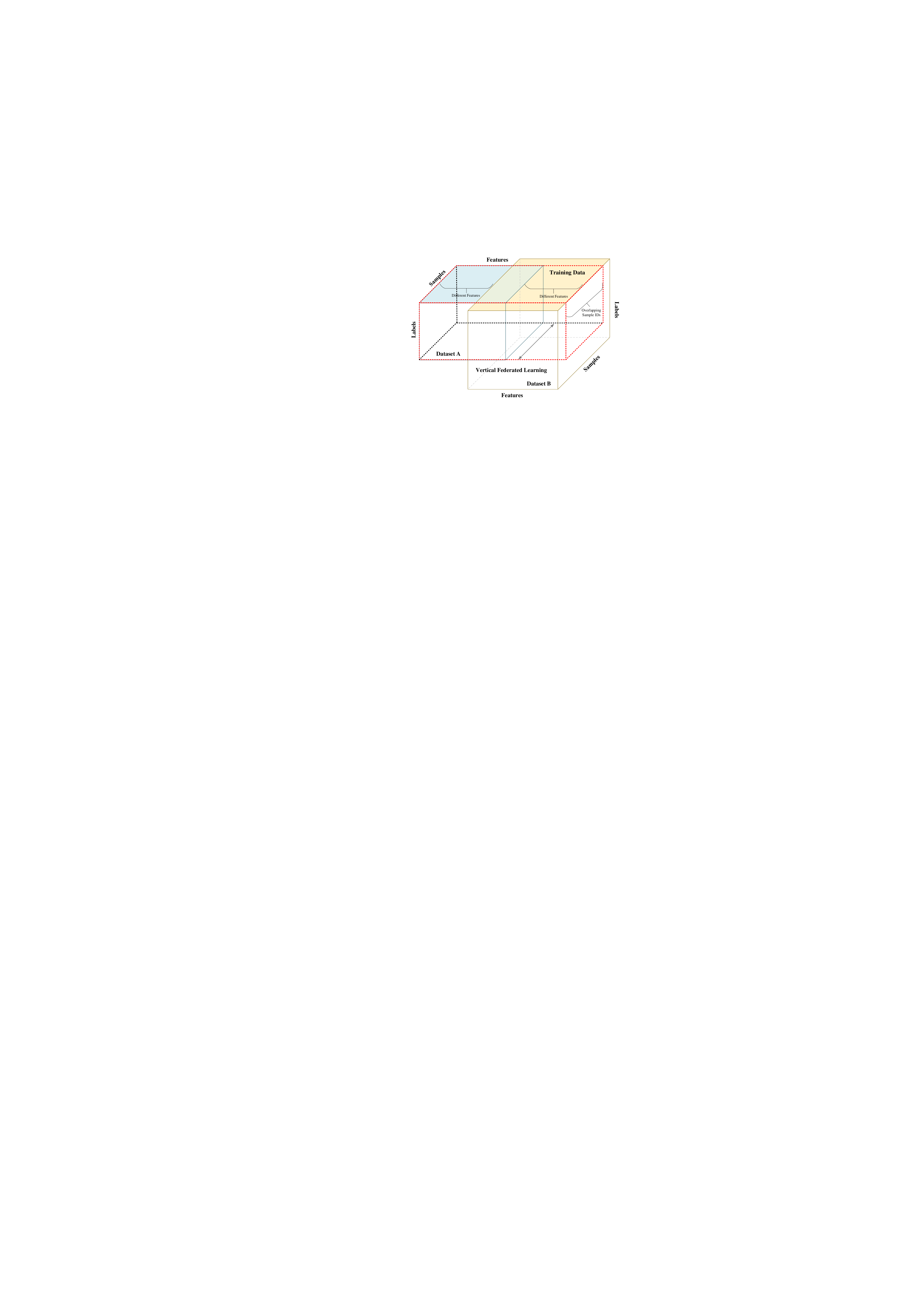}
		\caption{Illustration of vertical federated learning (VFL).}
		\label{fig-vertical}
	\end{figure}

\begin{figure}[htb]
		\centering
		\includegraphics[width=0.6\textwidth]{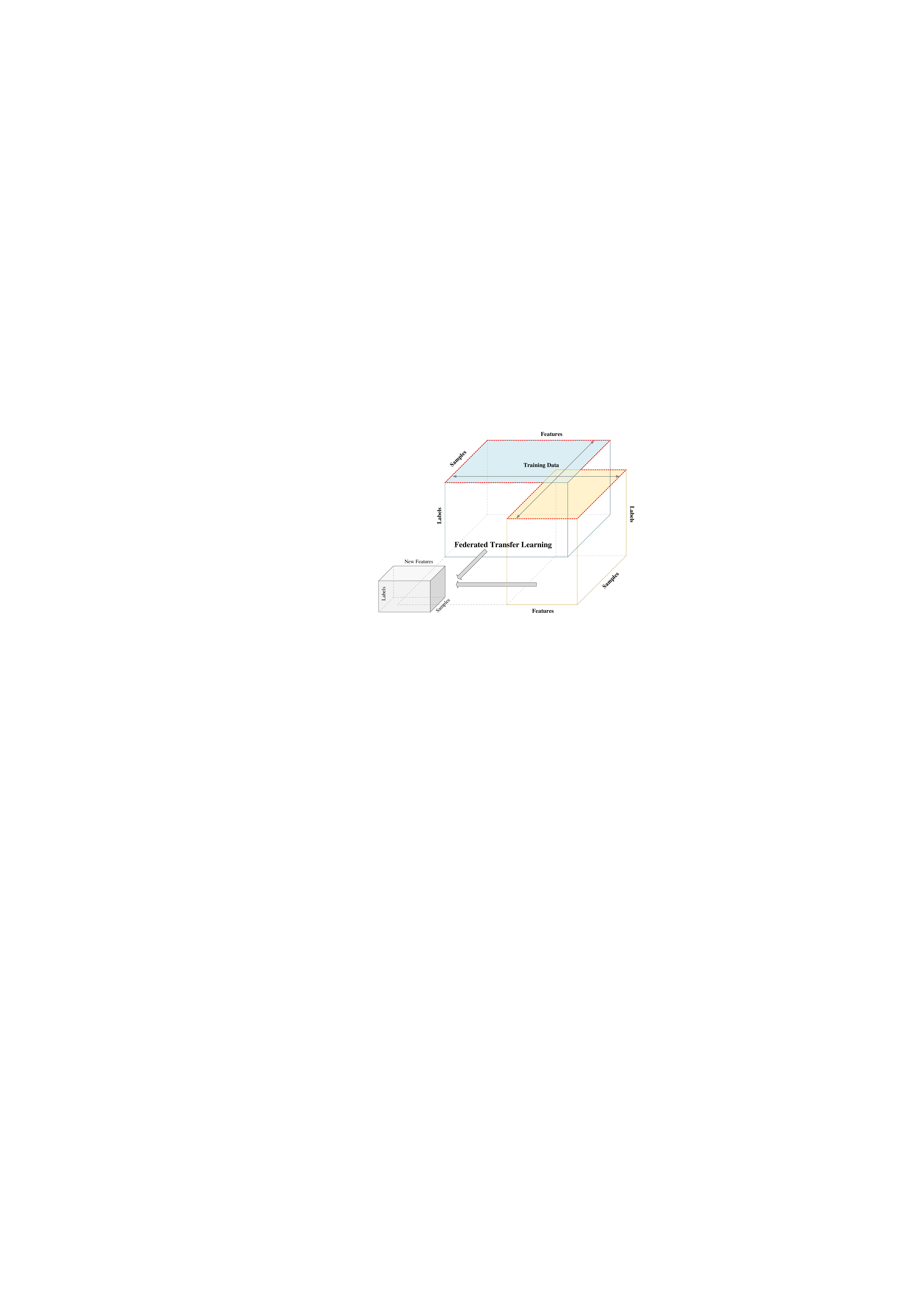}
		\caption{Illustration of federated transfer learning (FTL).}
		\label{fig-transfer}
	\end{figure}

	Based on the way data is partitioned within a feature and sample space, FL may be classified as HFL, VFL, or federated transfer learning (FTL)\cite{yang2019federated}. In Figure \ref{fig-horizontal}, Figure \ref{fig-vertical}, and Figure \ref{fig-transfer}, these three federated learning categories for a two-party scenario are illustrated. In order to define each category more clearly, some parameters are formalized. We suppose that the $i$-th participant has its own dataset $\mathcal{D} _i$. The dataset includes three types of data, \ie the feature space $\mathcal{X} _i$, the label space $\mathcal{Y} _i$ and the sample ID space $\mathcal{I} _i$. In particular, the feature space $\mathcal{X} _i$ is a high-dimensional abstraction of the variables within each pattern sample. Various features are used to characterize data held by the participant. All categories of association between input and task target are collected in the label space $\mathcal{Y} _i$. The sample ID space $\mathcal{I} _i$ is added in consideration of actual application requirements. The identification can facilitate the discovery of possible connections among different features of the same individual.
	
	HFL indicates the case in which participants have their dataset with a small sample overlap, while most of the data features are aligned. The word "horizontal" is derived from the term "horizontal partition". This is similar to the situation where data is horizontally partitioned inside the traditional tabular view of a database. As shown in Figure. \ref{fig-horizontal}, the training data of two participants with the aligned features is horizontally partitioned for HFL. A cuboid with a red border represents the training data required in learning. Especially, a row corresponds to complete data features collected from a sampling ID. Columns correspond to different sampling IDs. The overlapping part means there can be more than one participant sampling the same ID. In addition, HFL is also known as feature-aligned FL, sample-partitioned FL, or example-partitioned FL. Formally, the conditions for HFL can be summarized as
	$$
	\mathcal{X} _i=\mathcal{X} _j, \mathcal{Y} _i=\mathcal{Y} _j, \mathcal{I} _i\ne \mathcal{I} _j, \forall \mathcal{D} _i, \mathcal{D} _j, i\ne j,
	$$
	\noindent where $\mathcal{D} _i$  and $\mathcal{D} _j$ denote the datasets of participant $i$  and participant $j$ respectively. In both datasets, the feature space $\mathcal{X} $ and label space $\mathcal{Y} $ are assumed to be the same, but the sampling ID space $\mathcal{I} $ is assumed to be different. The objective of HFL is to increase the amount of data with similar features, while keeping the original data from being transmitted, thus improving the performance of the training model. Participants can still perform feature extraction and classification if new samples appear. HFL can be applied in various fields because it benefits from privacy protection and experience sharing \cite{DBLP:journals/corr/abs-1912-04977}. For example, regional hospitals may receive different patients, and the clinical manifestations of patients with the same disease are similar. It is imperative to protect the patient's privacy, so data about patients cannot be shared. HFL provides a good way to jointly build a ML model for identifying diseases between hospitals.
	
	VFL refers to the case where different participants with various targets usually have datasets that have different feature spaces, but those participants may serve a large number of common users. The heterogeneous feature spaces of distributed datasets can be used to build more general and accurate models without releasing the private data. The word “vertical” derives from the term “vertical partition”, which is also widely used in reference to the traditional tabular view. Different from HFL, the training data of each participant are divided vertically. Figure \ref{fig-vertical} shows an example of VFL in a two-party scenario. The important step in VFL is to align samples, \ie determine which samples are common to the participants. Although the features of the data are different, the sampled identity can be verified with the same ID. Therefore, VFL is also called sample-aligned FL or feature-partitioned FL. Multiple features are vertically divided into one or more columns. The common samples exposed to different participants can be marked by different labels. The formal definition of VFL's applicable scenario is given.
	$$
	\mathcal{X} _i\ne \mathcal{X} _j, \mathcal{Y} _i\ne \mathcal{Y} _j, \mathcal{I} _i=\mathcal{I} _j, \forall \mathcal{D} _i, \mathcal{D} _j, i\ne j,
	$$
	\noindent where $\mathcal{D}_i$ and $\mathcal{D}_i$ represent the dataset held by different participants, and the data feature space pair $\left( \mathcal{X} _i, \mathcal{X} _j \right)$ and label space pair $\left( \mathcal{Y} _i, \mathcal{Y} _j \right)$ are assumed to be different. The sample ID space $\mathcal{I} _i$  and $\mathcal{I} _j$ are assumed to be the same. It is the objective of VFL to collaborate in building a shared ML model by exploiting all features collected by each participant. The fusion and analysis of existing features can even infer new features. An example of the application of VFL is the evaluation of trust. Banks and e-commerce companies can create a ML model for trust evaluation for users. The credit card record held at the bank and the purchasing history held at the e-commerce company for the set of same users can be used as training data to improve the evaluation model.
	
	FTL applies to a more general case where the datasets of participants are not aligned with each other in terms of samples or features. FTL involves finding the invariant between a resource-rich source domain and a resource-scarce target domain, and exploiting that invariant to transfer knowledge. In comparison with traditional transfer learning \cite{5288526}, FTL focuses on privacy-preserving issues and addresses distributed challenges. An example of FTL is shown in Figure \ref{fig-transfer}. The training data required by FTL may include all data owned by multiply parties for comprehensive information extraction. In order to predict labels for unlabeled new samples, a prediction model is built using additional feature representations for mixed samples from participants A and B. More formally, FTL is applicable for the following scenarios:
	$$
	\mathcal{X} _i\ne \mathcal{X} _j, \mathcal{Y} _i\ne \mathcal{Y} _j, \mathcal{I} _i\ne \mathcal{I} _j, \forall \mathcal{D} _i, \mathcal{D} _j, i\ne j,
	$$
	In datasets $\mathcal{D} _i$  and $\mathcal{D} _j$ , there is no duplication or similarity in terms of features, labels and samples. The objective of FTL is to generate as accurate a label prediction as possible for newly incoming samples or unlabeled samples already present. Another benefit of FTL is that it is capable of overcoming the absence of data or labels. For example, a bank and an e-commerce company in two different countries want to build a shared ML model for user risk assessment. In light of geographical restrictions, the user groups of these two organizations have limited overlap. Due to the fact that businesses are different, only a small number of data features are the same. It is important in this case to introduce FTL to solve the problem of small unilateral data and fewer sample labels, and improve the model performance.
	
	
	
	\section{Reinforcement learning}
	\subsection{Reinforcement learning definition and basics}
	
	Generally, the field of ML includes supervised learning, unsupervised learning, RL, \etc \cite{li2017deep}. While supervised and unsupervised learning attempt to make the agent copy the data set, \ie learning from the pre-provided samples, RL is to make the agent gradually stronger in the interaction with the environment, \ie generating samples to learn by itself \cite{xu2018experience}. RL is a very hot research direction in the field of ML in recent years, which has made great progress in many applications, such as IoT \cite{mohammadi2017semisupervised,bu2019smart,9060882,8681129}, autonomous driving \cite{shalev2016safe,sallab2017deep}, and game design \cite{taylor2011teaching}. For example, the AlphaGo program developed by DeepMind is a good example to reflect the thinking of RL \cite{holcomb2018overview}. The agent gradually accumulates the intelligent judgment on the sub-environment of each move by playing game by game with different opponents, so as to continuously improve its level.\par 
	
\begin{figure}[htb]
		\centering
		\includegraphics[width=0.5\textwidth]{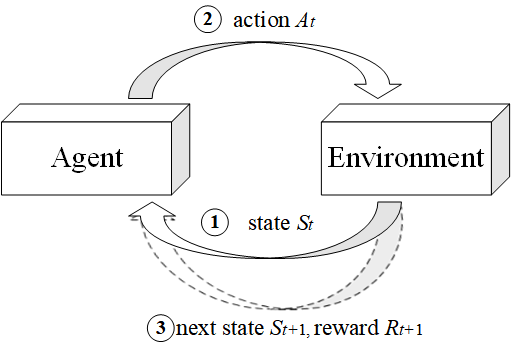}
		\caption{The agent-environment interaction of the basic reinforcement learning model.}
		\label{fig-RL-model}
	\end{figure}

	The RL problem can be defined as a model of the agent-environment interaction, which is represented in Figure \ref{fig-RL-model}. The basic model of RL contains several important concepts, \ie
	
	\begin{itemize}
		\item \textbf{Environment} and \textbf{agent}: Agents are a part of a RL model that exists in an external environment, such as the player in the environment of chess. Agents can improve their behavior by interacting with the environment. Specifically, they take a series of actions to the environment through a set of policies and expect to get a high payoff or achieve a certain goal. 
		\item \textbf{Time step}: The whole process of RL can be discretized into different time steps. At every time step, the environment and the agent interact accordingly. 
		\item \textbf{State}: The state reflects agents' observations of the environment. When agents take action, the state will also change. In other words, the environment will move to the next state.
		\item \textbf{Actions}: Agents can assess the environment, make decisions and finally take certain actions. These actions are imposed on the environment. 
		\item \textbf{Reward}: After receiving the action of the agent, the environment will give the agent the state of the current environment and the reward due to the previous action. Reward represents an assessment of the action taken by agents.
	\end{itemize}
	
	More formally, we assume that there are a series of time steps $t=\text{0,1,2,}...$ in a basic RL model. At a certain time step $t$, the agent will receive a state signal $S_t$ of the environment. In each step, the agent will select one of the actions allowed by the state to take an action $A_t$. After the environment receives the action signal $A_t$, the environment will feed back to the agent the corresponding status signal $S_{t+1}$ at the next step $t+1$ and the immediate reward $R_{t+1}$. The set of all possible states, \ie the state space, is denoted as $\mathcal{S}$. Similarly, the action space is denoted as $\mathcal{A}$. Since our goal is to maximize the total reward, we can quantify this total reward, usually referred to as return with
	$$
	G_t=R_{t+1}+R_{t+2}+...+R_T,
	$$
	\noindent where $T$ is the last step, i.e., $S_T$ as the termination state. An \textbf{episode} is completed when the agent completes the termination action. 
	
	In addition to this type of episodic task, there is another type of task that does not have a termination state, in other words, it can in principle run forever. This type of task is called a continuing task. For continuous tasks, since there is no termination state, the above definition of return may be divergent. Thus, another way to calculate return is introduced, which is called discounted return, \ie
	$$
	G_t=R_{t+1}+\gamma R_{t+2}+\gamma ^2R_{t+3}+...=\sum_{k=0}^{\infty}{\gamma ^k}R_{t+k+1},
	$$
	\noindent where the discount factor $\gamma$ satisfies $0\leqslant \gamma \leqslant 1$. When $\gamma = 1$, the agent can obtain the full value of all future steps, while when $\gamma =0$, the agent can only see the current reward. As $\gamma$ changes from $0$ to $1$, the agent will gradually become forward-looking, looking not only at current interests, but also for its own future. 
	
	The value function is the agent's prediction of future rewards, which is used to evaluate the quality of the state and select actions. The difference between the value function and rewards is that the latter is defined as evaluating an immediate sense for interaction while the former is defined as the average return of actions over a long period of time. In other words, the value function of the current state $S_t=s$ is its long-term expected return. There are two significant value functions in the field of RL, \ie state value function $V_{\pi}\left( s \right)$ and action value function $Q_{\pi}\left( s,a \right)$. The function $V_{\pi}\left( s \right)$ represents the expected return obtained if the agent continues to follow strategy $\pi$ all the time after reaching a certain state $S_t$, while the function $Q_{\pi}\left( s,a \right)$ represents the expected return obtained if action $A_t=a$ is taken after reaching the current state $S_t=s$ and the following actions are taken according to the strategy $\pi$. The two functions are specifically defined as follows, \ie

	$$
	V_{\pi}\left( s \right) =\mathbb{E}_{\pi}\left[ G_t|S_t=s \right], \forall s\in\mathcal{S}
	$$
	$$
	Q_{\pi}\left( s,a \right) =\mathbb{E}_{\pi}\left[ G_t|S_t=s,A_t=a \right], \forall s\in\mathcal{S}, a\in\mathcal{A}.
	$$

	The results of RL are action decisions, called as the policy. The policy gives agents the action $a$ that should be taken for each state $s$. It is noted as =$\pi \left( A_t=a|S_t=s \right)$, which represents the probability of taking action $A_t=a$ in state $S_t=s$. The goal of RL is to learn the optimal policy that can maximize the value function by interacting with the environment. Our purpose is not to get the maximum reward after a single action in the short term, but to get more reward in the long term. Therefore, the policy can be figured out as, 
	$$
	\pi^{*} =arg\underset{\pi}{\max}V_{\pi}\left( s \right), \forall s\in\mathcal{S}.
	$$
	
	\subsection{Categories of reinforcement learning}
\begin{figure}[htb]
		\centering
		\includegraphics[width=0.5\textwidth]{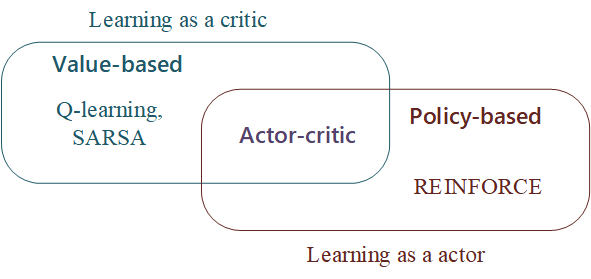}
		\caption{The categories and representative algorithms of reinforcement learning.}
		\label{fig-RL-categories}
	\end{figure}

	In RL, there are several categories of algorithms. One is value-based and the other is policy-based. In addition, there is also an actor-critic algorithm that can be obtained by combining the two, as shown in Figure \ref{fig-RL-categories}.
	
	\subsubsection{Value-based methods}
	
	Recursively expand the formulas of the action value function, the corresponding Bellman equation is obtained, which describes the recursive relationship between the value function of the current state and subsequent state. The recursive expansion formula of the action value function $Q_{\pi}\left( s,a \right)$ is
	$$
	Q_{\pi}\left( s,a \right) = \sum_{s',r}{p\left( s^{'},r|s,a \right)}\left[ r+\gamma \sum_{a'}{\pi \left( a^{'}|s^{'} \right) Q_{\pi}\left( s^{'},a^{'} \right)} \right],
	$$
	\noindent where the function $p\left( s^{'},r|s,a \right) =Pr\left\{ S_t=s',R_t=r|S_{t-1}=s,A_{t-1}=a \right\} 
	$ defines the trajectory probability to characterize the environment's dynamics. $R_{t}=r$ indicates the reward obtained by the agent taking action $A_{t-1}=a$ in state $S_{t-1}=s$. Besides, $S_{t}=s'$ and $A_{t}=a'$ respectively represent the state and the action taken by the agent at the next moment $t$.
	
	In the value-based algorithms, the above value function $Q_{\pi}\left( s,a \right)$ is calculated iteratively, and the strategy is then improved based on this value function. If the value of every action in a given state is known, the agent can select an action to perform. In this way, if the optimal  $Q_{\pi}\left( s,a=a^* \right) $ can be figured out, the best action $a^*$ will be found. There are many classical value-based algorithms, including Q-learning \cite{watkins1992q}, State–Action–Reward–State–Action (SARSA) \cite{thorpe1997vehicle}, \etc.
	
	
	
	Q-learning is a typical widely-used value-based RL algorithm. It is also a model-free algorithm, which means that it does not need to know the model of the environment but directly estimates the Q value of each executed action in each encountered state through interacting with the environment \cite{watkins1992q}. Then, the optimal strategy is formulated by selecting the action with the highest Q value in each state. This strategy maximizes the expected return for all subsequent actions from the current state. The most important part of Q-learning is the update of Q value. It uses a table, \ie Q-table, to store all Q value functions. Q-table uses state as row and action as column. Each $(s,a)$ pair corresponds to a Q value, i.e. $Q(s,a)$, in the Q-table, which is updated as follows, 
	$$
	Q\left( s,a \right) \gets Q\left( s,a \right) +\alpha \left[ r+\gamma \underset{a'}{\max}Q\left( s^{'},a^{'} \right) -Q\left( s,a \right) \right] 
	$$
	\noindent where $r$ is the reward given by taking action $a$ under state $s$ at the current time step. $s'$ and $a'$ indicate the state and the action taken by the agent at the next time step respectively. $\alpha$ is the learning rate to determine how much error needs to be learned, and $\gamma$ is the attenuation of future reward. If the agent continuously accesses all state-action pairs, the Q-learning algorithm will converge to the optimal Q function. Q-learning is suitable for simple problems, \ie small state space, or a small number of actions. It has high data utilization and stable convergence.
	
	\subsubsection{Policy-based methods}
	
	The above value-based method is an indirect approach to policy selection, and has trouble handling an infinite number of actions. Therefore, we want to be able to model the policy directly. Different from the value-based method, the policy-based algorithm does not need to estimate the value function, but directly fits the policy function, updates the policy parameters through training, and directly generates the best policy. In policy-based methods, we input a state and output the corresponding action directly, rather than the value $V\left( s \right)$ or Q value $Q\left( s,a \right)$ of the state. One of the most representative algorithms is strategy gradient, which is also the most basic policy-based algorithm. 
	
	Policy gradient chooses to optimize the policy directly and update the parameters of the policy network by calculating the gradient of expected reward \cite{pmlr-v32-silver14}. Therefore, its objective function $J\left( \theta \right) $ is directly designed as expected cumulative rewards, \ie
	$$
	J\left( \theta \right) =\mathbb{E}_{\tau ~\_{\theta}\left( \tau \right)}\left[ r\left( \tau \right) \right] =\int_{\tau ~\pi \left( \tau \right)}{\begin{array}{c}
			r\left( \tau \right) \pi _{\theta}\left( \tau \right) d\tau\\
	\end{array}}.
	$$
	
	By taking the derivative of $J\left( \theta \right) $, we get
	
	$$
	\nabla _{\theta}J\left( \theta \right) =\mathbb{E}_{\tau ~\pi _{\theta}\left( \tau \right)}\left[ \sum_{t=1}^T{\nabla _{\theta}\log\pi _{\theta}\left( A_t|S_t \right) \sum_{t=1}^T{r\left( S_t,A_t \right)}} \right]. 
	$$
	
	The above formula consists of two parts. One is $\sum_{t=1}^T{\nabla _{\theta}\log\pi _{\theta}\left( A_t|S_t \right)}$ which denotes the probability of the gradient in the current trace. The other is $\sum_{t=1}^T{r\left( S_t,A_t \right)}$ which represents the return of the current trace. Since the return is total rewards  and can only be obtained after one episode, the policy gradient algorithm can only be updated for each episode, not for each time step.
	
	The expected value can be expressed in a variety of ways, corresponding to different ways of calculating the loss function. The advantage of the strategy gradient algorithm is that it can be applied in the continuous action space. In addition, the change of the action probability is smoother, and the convergence is better guaranteed. 
	
	REINFORCE algorithm is a classic policy gradient algorithm \cite{williams1992simple}. Since the expected value of the cumulative reward cannot be calculated directly, the Monte Carlo method is applied to approximate the average value of multiple samples. REINFORCE updates the unbiased estimate of the gradient by using Monte Carlo sampling. Each sampling generates a trajectory, which runs iteratively. After obtaining a large number of trajectories, the cumulative reward can be calculated by using certain transformations and approximations as the loss function for gradient update. However, the variance of this algorithm is large since it needs to interact with the environment until the terminate state. The reward for each interaction is a random variable, so each variance will add up when the variance is calculated.
	In particular, the REINFORCE algorithm has three steps:
	\begin{itemize}
		\item Step 1: sample $\tau_i$ from $\pi _{\theta}\left( A_t|S_t \right)$ 
		\item Step 2: $\nabla _{\theta}J\left( \theta \right) \approx \sum_i{\left[ \sum_{t=1}^T{\nabla _{\theta}\log\pi _{\theta}\left( A_{t}^{i}|S_{t}^{i} \right) \sum_{t=1}^T{r\left( S_{t}^{i},A_{t}^{i} \right)}} \right]}$
		\item Step 3: $theta \gets \theta +\alpha \nabla _{\theta}J\left( \theta \right)$
	\end{itemize}
	
	
	The two algorithms, value-based and policy-based methods, both have their own characteristics and disadvantages. Firstly, the disadvantages of the value-based methods are that the output of the action cannot be obtained directly, and it is difficult to extend to the continuous action space. The value-based methods can also lead to the problem of high bias, \ie it is difficult to eliminate the error between the estimated value function and the actual value function. For the policy-based methods, a large number of trajectories must be sampled, and the difference between each trajectory may be huge. As a result, high variance and large gradient noise are introduced. It leads to the instability of training and the difficulty of policy convergence.
	
	\subsubsection{Actor-critic methods}
	
	The actor-critic architecture combines the characteristics of the value-based and policy-based algorithms, and to a certain extent solves their respective weaknesses, as well as the contradictions between high variance and high bias. The constructed agent can not only directly output policies, but also evaluate the performance of the current policies through the value function. Specifically, the actor-critic architecture consists of an actor which is responsible for generating the policy and a critic to evaluate this policy. When the actor is performing, the critic should evaluate its performance, both of which are constantly being updated \cite{konda2000actor}. This complementary training is generally more effective than a policy-based method or value-based method.
	
	In specific, the input of actor is state $S_t$, and the output is action $A_t$. The role of actor is to approximate the policy model $\pi _{\theta}\left( A_t|S_t \right) $. Critic uses the value function $Q$ as the output to evaluate the value of the policy, and this Q value $Q\left( S_t,A_t \right) $ can be directly applied to calculate the loss function of actor. The gradient of the expected revenue function $J\left(\theta \right)$ in the action-critic framework is developed from the basic policy gradient algorithm. The policy gradient algorithm can only implement the update of each episode, and it is difficult to accurately feedback the reward. Therefore, it has poor training efficiency. Instead, the actor-critic algorithm replaces $\sum_{t=1}^T{r\left( S_{t}^{i},A_{t}^{i} \right)}$ with $Q\left( S_t,A_t \right) $ to evaluate the expected returns of state-action tuple $\left\{S_t,A_t \right\}$ in the current time step $t$. The gradient of $J\left(\theta \right)$ can be expressed as
	
	$$
	\nabla _{\theta}J\left( \theta \right) =\mathbb{E}_{\tau ~\pi _{\theta}\left( \tau \right)}\left[ \sum_{t=1}^T{\nabla _{\theta}\log\pi _{\theta}\left( A_t|S_t \right) Q\left( S_t,A_t \right)} \right].
	$$

	\subsection{Deep reinforcement learning}
	
	With the continuous expansion of the application of deep learning, its wave also swept into the RL field, resulting in Deep Reinforcement Learning (DRL), \ie using a multi-layer deep neural network to approximate value function or policy function in the RL algorithm \cite{henderson2018deep,9277511}. DRL mainly solves the curse-of-dimensionality problem in real-world RL applications with large or continuous state and/or action space, where the traditional tabular RL algorithms cannot store and extract a large amount of feature information \cite{li2017deep,8802256}. 
	
	Q-learning, as a very classical algorithm in RL, is a good example to understand the purpose of DRL. The big issue with Q-learning falls into the tabular method, which means that when state and action spaces are very large, it cannot build a very large Q table to store a large number of Q values \cite{ohnishi2019constrained}. Besides, it counts and iterates Q values based on past states. Therefore, on the one hand, the applicable state and action space of Q-learning is very small. On the other hand, if a state never appears, Q-learning cannot deal with it \cite{peng1994incremental}. In other words, Q-learning has no prediction ability and generalization ability at this point. 
	
	In order to make Q-learning with prediction ability, considering that neural network can extract feature information well, Deep Q Network (DQN) is proposed by applying deep neural network to simulate Q value function. In specific, DQN is the continuation of Q-learning algorithm in continuous or large state space to approximate Q value function by replacing Q table with neural networks \cite{mnih2015human}. 
	
	In addition to the value-based DRL algorithm such as DQN, we summarize a variety of classical DRL algorithms according to algorithm types by referring to some DRL related surveys \cite{lei2020deep} in Table \ref{drl_algorithms_table}, including not only the policy-based and actor-critic DRL algorithms, but also the advanced DRL algorithms of Partially Observable Markov Decision Process (POMDP) and multi-agents.

	\begin{table}[htb]
		\centering
		\caption{Taxonomy of representative algorithms for DRL.}
		\label{drl_algorithms_table}
		\renewcommand\arraystretch{1.1}
		\begin{tabular}{|p{2cm}<{\centering}|p{2cm}<{\centering}|p{10cm}|}
			\hline
			\multicolumn{2}{|c|}{\textbf{Types}} & \textbf{Representative algorithms}                                                                                                                                                                                                                                                                                                                                                         \\ \hline
			\multicolumn{2}{|c|}{Value-based}         & \begin{tabular}[c]{@{}l@{}}Deep Q-Network (DQN) \cite{mnih2015human}, Double Deep Q-Network (DDQN) \cite{van2016deep}, \\ DDQN with proportional prioritization \cite{schaul2015prioritized}\end{tabular}                                                                                                                                                                                                                                                       \\ \hline
			\multicolumn{2}{|c|}{Policy-based}        & \begin{tabular}[c]{@{}l@{}}REINFORCE \cite{williams1992simple}, Q-prop \cite{gu2016q}\end{tabular}                                                                                                                                                                                                                                                                                                                                                                 \\ \hline
			\multicolumn{2}{|c|}{Actor-critic}        & \begin{tabular}[c]{@{}l@{}}Soft Actor-Critic (SAC) \cite{haarnoja2018soft}, Asynchronous Advantage Actor Critic (A3C) \cite{mnih2016asynchronous}, \\ Deep Deterministic Policy Gradient (DDPG) \cite{lillicrap2015continuous}, \\ Distributed Distributional Deep Deterministic Policy Radients (D4PG) \cite{barth2018distributed}, \\ Twin Delayed Deep Deterministic (TD3) \cite{fujimoto2018addressing}, \\ Trust Region Policy Optimization (TRPO) \cite{schulman2015trust}, \\ Proximal Policy Optimization (PPO) \cite{schulman2017proximal}\end{tabular} \\ \hline
			\multirow{6}{*}{Advanced}  & POMDP        & \begin{tabular}[c]{@{}l@{}}Deep Belief Q-Network (DBQN) \cite{zhu2017improving}, \\ Deep Recurrent Q-Network (DRQN) \cite{hausknecht2015deep}, \\ Recurrent Deterministic Policy Gradients (RDPG) \cite{heess2015memory} \end{tabular}                                                                                                                                                                                                    \\ \cline{2-3} 
			& Multi-agents & \begin{tabular}[c]{@{}l@{}}Multi-Agent Importance Sampling (MAIS) \cite{foerster2017stabilising}, \\ Coordinated Multi-agent DQN \cite{van2016coordinated}, \\ Multi-agent Fingerprints (MAF) \cite{foerster2017stabilising}, \\ Counterfactual Multiagent Policy Gradient (COMAPG) \cite{foerster2018counterfactual}, \\ Multi-Agent DDPG (MADDPG) \cite{lowe2017multi}\end{tabular}                                                                                                        \\ \hline
		\end{tabular}
	\end{table}
	
	\section{Federated Reinforcement Learning}
	In this section, the detailed background and categories of FRL will be discussed. 
	
	\subsection{Federated reinforcement learning background}
	
	Despite the excellent performance that RL and DRL have achieved in many areas, they still face several important technical and non-technical challenges in solving real-world problems. The successful application of FL in supervised learning tasks arouses interest in exploiting similar ideas in RL, \ie FRL. On the other hand, although FL is useful in some specific situations, it fails to deal with cooperative control and optimal decision-making in dynamic environments \cite{9084352}. FRL not only provides the experience for agents to learn to make good decisions in an unknown environment, but also ensures that the privately collected data during the agent's exploration does not have to be shared with others. A forward-looking and interesting research direction is how to conduct RL under the premise of protecting privacy. Therefore, it is proposed to use FL framework to enhance the security of RL and define FRL as a security-enhanced distributed RL framework to accelerate the learning process, protect agent privacy and handle Not Independent and Identically Distributed (Non-IID) data \cite{yang2019federated}. Apart from improving the security and privacy of RL, we believe that FRL has a wider and larger potential in helping RL to achieve better performance in various aspects, which will be elaborated in the following subsections.
	
\begin{figure}[htb]
		\centering
		\includegraphics[width=1.0\textwidth]{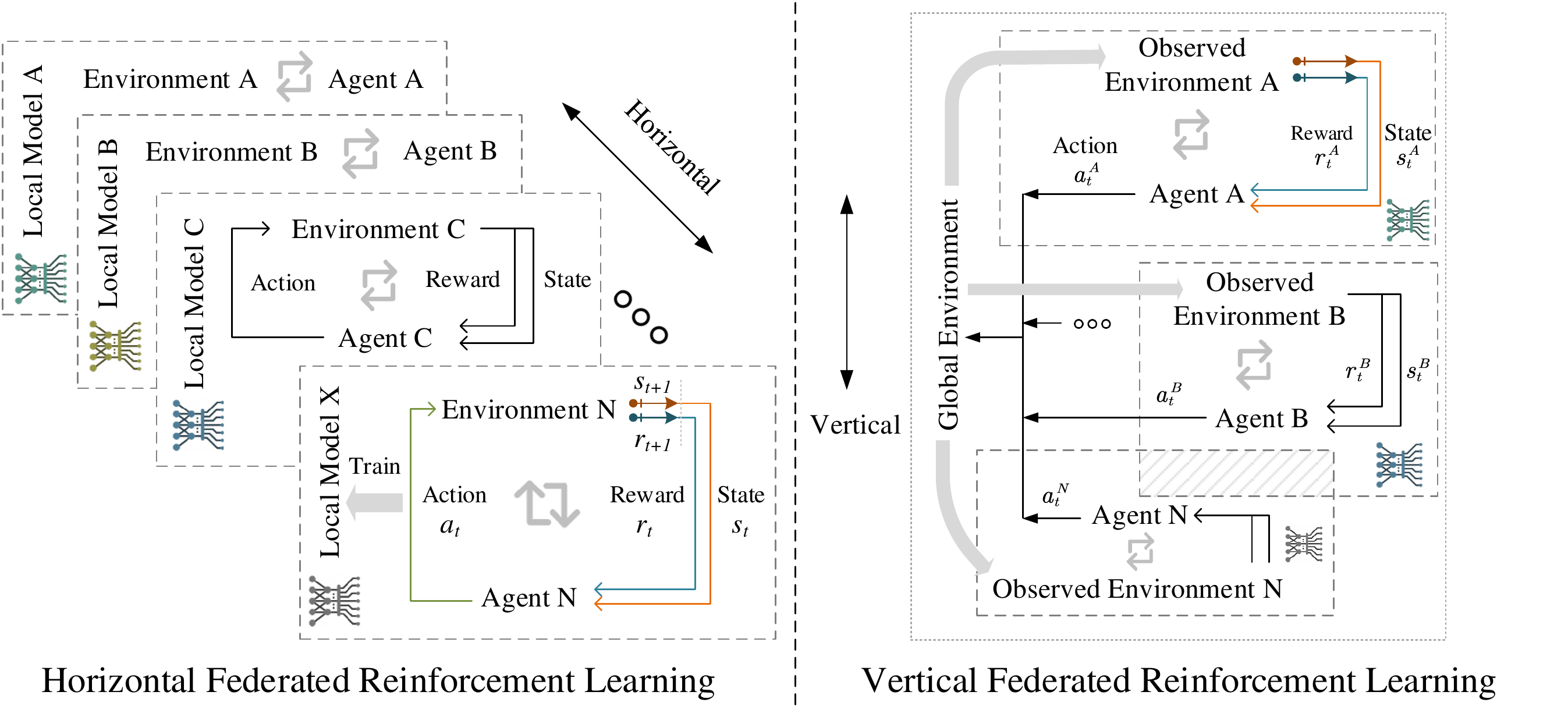}
		\caption{Comparison of horizontal federated reinforcement learning (HFRL) and vertical federated reinforcement learning (VFRL).}
		\label{fig-hfrl_vfrl}
	\end{figure}

	In order to facilitate understanding and maintain consistency with FL, FRL is divided into two categories depending on environment partition \cite{bookfederatedlearning}, \ie HFRL and VFRL. Figure \ref{fig-hfrl_vfrl} gives the comparison between HFRL and VFRL. In HFRL, the environment that each agent interacts with is independent of the others, while the state space and action space of different agents are aligned to solve similar problems. The action of each agent only affects its own environment and results in corresponding rewards. As an agent can hardly explore all states of its environment, multiple agents interacting with their own copy of the environment can accelerate training and improve model performance by sharing experience. Therefore, horizontal agents use server-client model or peer-to-peer model to transmit and exchange the gradients or parameters of their policy models (actors) and/or value function models (critics). In VFRL, multiple agents interact with the same global environment, but each can only observe limited state information in the scope of its view. Agents can perform different actions depending on the observed environment and receive local reward or even no reward. Based on the actual scenario, there may be some observation overlap between agents. In addition, all agents' actions affect the global environment dynamics and total rewards. As opposed to the horizontal arrangement of independent environments in HFRL, the vertical arrangement of observations in VFRL poses a more complex problem and is less studied in the existing literature.
	
	\subsection{Horizontal federated reinforcement learning}
	HFRL can be applied in scenarios in which the agents may be distributed geographically, but they face similar decision-making tasks and have very little interaction with each other in the observed environments. Each participating agent independently executes decision-making actions based on the current state of environment and obtains positive or negative rewards for evaluation. Since the environment explored by one agent is limited and each agent is unwilling to share the collected data, multiple agents try to train the policy and/or value model together to improve model performance and increase learning efficiency. The purpose of HFRL is to alleviate the sample-efficiency problem in RL, and help each agent quickly obtain the optimal policy which can maximize the expected cumulative reward for specific tasks, while considering privacy protection.
	
	
	In the HFRL problem, the environment, state space, and action space can replace the data set, feature space, and label space of basic FL. More formally, we assume that $N$ agents $\left\{ \mathcal{F} _i \right\} _{i=1}^{N}$  can observe the environment $ \left\{ \mathcal{E} _i \right\} _{i=1}^{N}$ within their field of vision. $\mathcal{G}$ denotes the collection of all environments. The environment $\mathcal{E}_i$ where the $i$-th agent is located has a similar model, \ie state transition probability and reward function compared to other environments. Note that the environment $ \mathcal{E}_i$ is independent of the other environments, in that the state transition and reward model of $ \mathcal{E}_i$ do not depend on the states and actions of the other environments. Each agent $ \mathcal{F}_i$ interacts with its own environment $ \mathcal{E}_i$ to learn an optimal policy. Therefore, the conditions for HFRL are presented as follows, \ie
	
	$$
	\mathcal{S} _i=\mathcal{S} _j, \mathcal{A} _i=\mathcal{A} _j, \mathcal{E} _i\ne \mathcal{E} _j, \forall i,j\in \left\{ \text{1,2,...,}N \right\}, \mathcal{E} _i,\mathcal{E} _j \in \mathcal{G} ,i\ne j,
	$$
	
	\noindent where $\mathcal{S} _i$ and $\mathcal{S} _j$ denote the similar state space encountered by the $i$-th agent and $j$-th agent, respectively. $\mathcal{A} _i$ and $\mathcal{A} _j$ denote the similar action space of the $i$-th agent and $j$-th agent, respectively The observed environment $\mathcal{E} _i$ and $\mathcal{E} _i$ are two different environments that are assumed to be independent and ideally identically distributed. 
	
\begin{figure}[htb]
		\centering
		\includegraphics[width=0.6\textwidth]{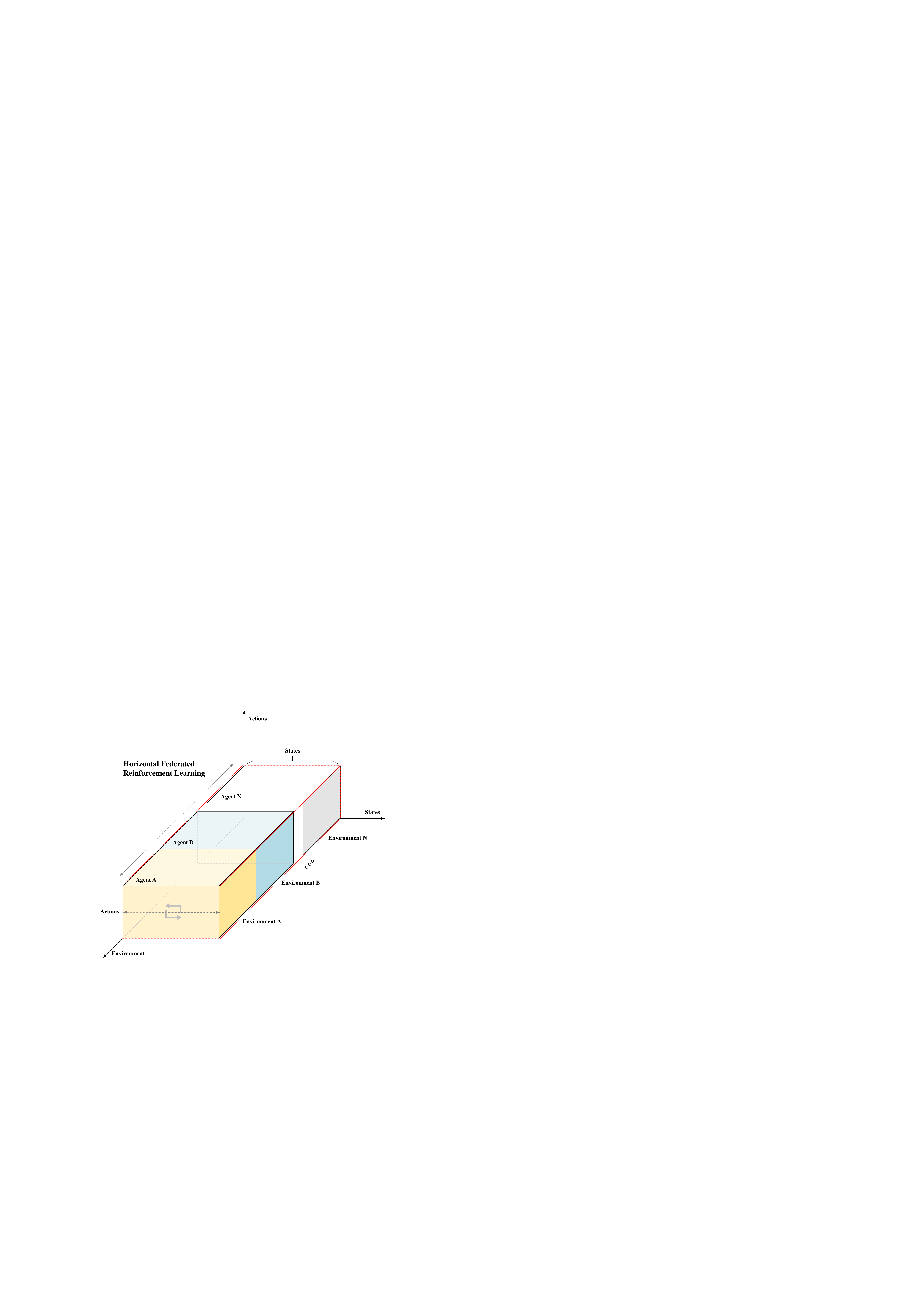}
		\caption{Illustration of horizontal federated reinforcement learning (HFRL).}
		\label{fig-hfrl-arch}
	\end{figure}

	Figure \ref{fig-hfrl-arch} shows the HFRL in graphic form. Each agent is represented by a cuboid. The axes of the cuboid denote three dimensions of information, \ie the environment, state space, and action space. We can intuitively see that all environments are arranged horizontally, and multiple agents have aligned state and action spaces. In other words, each agent explores independently in its respective  environment, and needs to obtain optimal strategies for similar tasks. In HFRL, agents share their experiences by exchanging masked models to enhance sample efficiency and accelerate the learning process. 
	
	A typical example of HFRL is the autonomous driving system in IoV. As vehicles drive on roads throughout the city and country, they can collect various environmental information and train the autonomous driving models locally. Due to driving regulations, weather conditions, driving routes, and other factors, one vehicle cannot be exposed to every possible situation in the environment. Moreover, the vehicles have basically the same operations, including braking, acceleration, steering, \etc. Therefore, vehicles driving on different roads, different cities, or even different countries could share their learned experience with each other by FRL without revealing their driving data according to the premise of privacy protection. In this case, even if other vehicles have never encountered a situation, they can still perform the best action by using the shared model. The exploration of multiple vehicles together also creates an increased chance of learning rare cases to ensure the reliability of the model.
	
\begin{figure}[htb]
		\centering
		\includegraphics[width=0.8\textwidth]{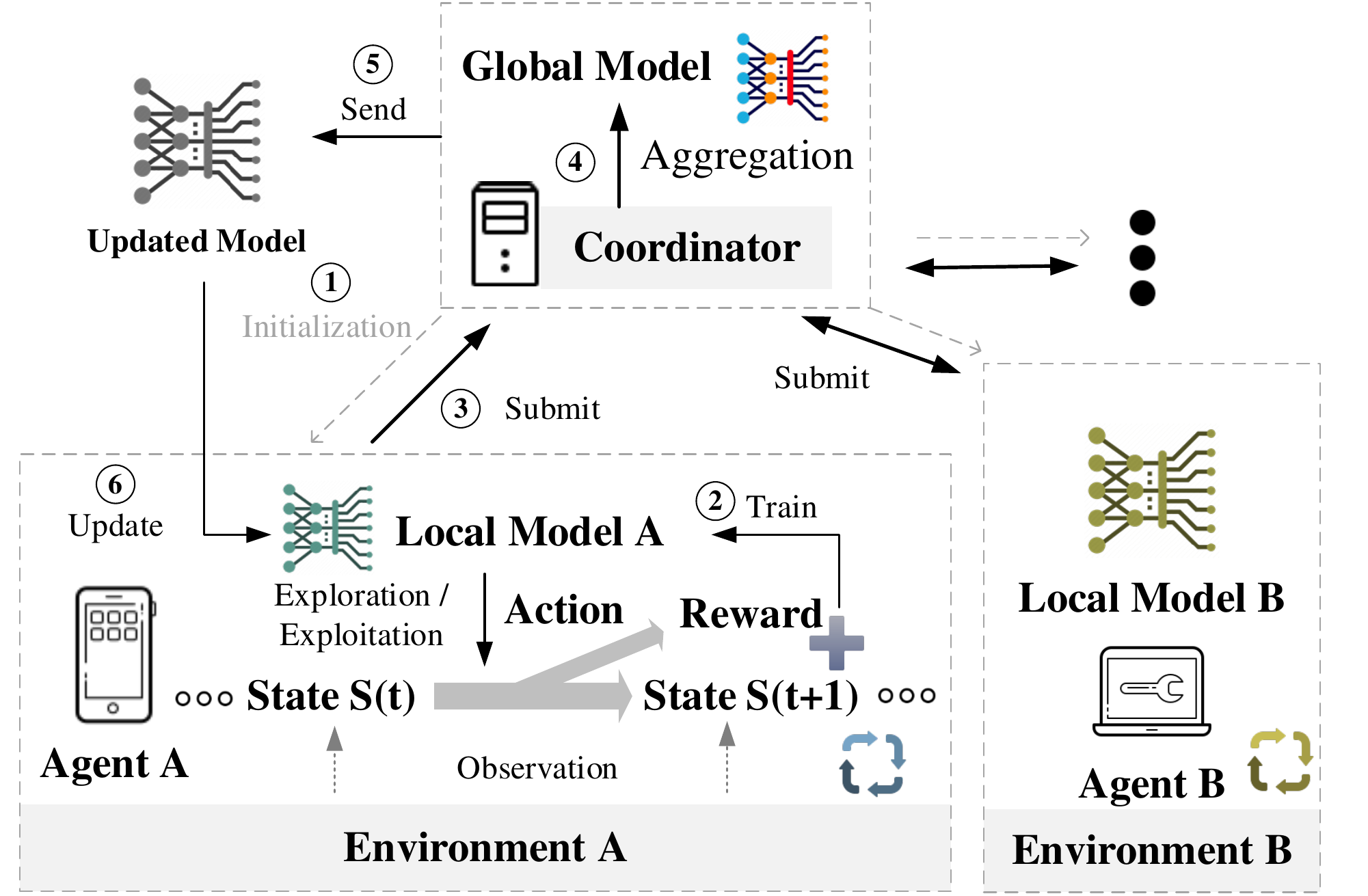}
		\caption{An example of horizontal federated reinforcement learning (HFRL) architecture.}
		\label{fig-hfrl}
	\end{figure}

	For a better understanding of HFRL, Figure \ref{fig-hfrl} shows an example of HFRL architecture using the server-client model. The coordinator is responsible for establishing encrypted communication with agents and implementing aggregation of shared models. The multiple parallel agents may be composed of heterogeneous equipment (\eg IoT devices, smart phone and computers, \etc.) and distributed geographically. It is worth noting that there is no specific requirement for the number of agents, and agents are free to choose to join or leave. The basic procedure for conducting HFRL can be summarized as follows.
	\begin{itemize}
		\item Step 1: The initialization/join process can be divided into two cases, one is when the agent has no model locally, and the other is when the agent has a model locally. For the first case, the agent can directly download the shared global model from a coordinator. For the second case, the agent needs to confirm the model type and parameters with the central coordinator.
		\item Step 2: Each agent independently observes the state of the environment and determines the private  strategy based on the local model. The selected action is evaluated by the next state and received reward. 
		All agents train respective models in state-action-reward-state (SARS) cycles.
		\item Step 3: Local model parameters are encrypted and transmitted to the coordinator. Agents may submit local models at any time as long as the trigger conditions are met.
		\item Step 4: The coordinator conducts the specific aggregation algorithm to evolve the global federated model. Actually, there is no need to wait for submissions from all agents, and appropriate aggregation conditions can be formulated depending on communication resources.
		\item Step 5: The coordinator sends back the aggregated model to the agents. 
		\item Step 6: The agents improve their respective models by fusing the federated model.
	\end{itemize}
	
	Following the above architecture and process, applications suitable for HFRL should meet the following characteristics. First, agents have similar tasks to make decisions under dynamic environments. Different from the FL setting, the goal of the HFRL-based application is to find the optimal strategy to maximize reward in the future. For the agent to accomplish the task requirements, the optimal strategy directs them to perform certain actions, such as control, scheduling, navigation, \etc. Second, distributed agents maintain independent observations. Each agent can only observe the environment within its field of view, but does not ensure that the collected data follows the same distribution. Third, it is important to protect the data that each agent collects and explores. Agents are presumed to be honest but curious, \ie they honestly follow the learning mechanism but are curious about private information held by other agents. Due to this, the data used for training is only stored at the owner and is not transferred to the coordinator. HFRL provides an implementation method for sharing experiences under the constraints of privacy protection. Additionally, various reasons limit the agent's ability to explore the environment in a balanced manner. Participating agents may include heterogeneous devices. The amount of data collected by each agent is unbalanced due to mobility, observation, energy and other factors. However, all participants have sufficient computing, storage, and communication capabilities. These capabilities assist the agent in completing model training, merging, and other basic processes. Finally, the environment observed by a agent may change dynamically, causing differences in data distribution. The participating agents need to update the model in time to quickly adapt to environmental changes and construct a personalized local model.
	
	In existing RL studies, some applications that meet the above characteristics can be classified as HFRL. Nadiger \etal \cite{8791693} presents a typical HFRL architecture, which includes the grouping policy, the learning policy, and the federation policy. In this work, RL is used to show the applicability  of granular personalization and FL is used to reduce training time. To demonstrate the effectiveness of the proposed architecture, a non-player character in the Atari game Pong is implemented and evaluated. In the study from Liu \etal \cite{DBLP:journals/corr/abs-1901-06455}, the authors propose the Lifelong Federated Reinforcement Learning (LFRL) for navigation in cloud robotic systems. It enables the robot to learn efficiently in a new environment and use prior knowledge to quickly adapt to the changes in the environment. Each robot trains a local model according to its own navigation task, and the centralized cloud server implements a knowledge fusion algorithm for upgrading a shared model. In considering that the local model and the shared model might have different network structures, this paper proposes to apply transfer learning to improve the performance and efficiency of the shared model. Further, researchers also focus on HFRL-based applications in the IoT due to the high demand for privacy protection. Ren \etal \cite{8728285} suggest deploying the FL architecture between edge nodes and IoT devices for computation offloading tasks. IoT devices can download RL model from edge nodes and train the local model using own data, including the remained energy resources and the workload of IoT device, \etc. The edge node aggregates the updated private model into the shared model. Although this method considers privacy protection issues, it requires further evaluation regarding the cost of communication resources by the model exchange. In addition, the work \cite{9062302} proposes a federated deep-reinforcement-learning-based framework (FADE) for edge caching. Edge devices, including Base Stations (BSs), can cooperatively learn a predictive model using the first round of training parameters for local learning, and then upload the local parameters tuned to the next round of global training. By keeping the training on local devices, the FADE can enable fast training and decouple the learning process between the cloud and data owner in a distributed-centralized manner. More HFRL-based applications will be classified and summarized in the next section.
	
	Prior to HFRL, a variety of distributed RL algorithms have been extensively investigated, which are closely related to HFRL. In general, distributed RL algorithms can be divided into two types: synchronized and asynchronous. In synchronous RL algorithms, such as Sync-Opt Synchronous Stochastic Optimization (Sync-Opt) \cite{DBLP:journals/corr/ChenMBJ16} and Parallel Advantage Actor Critic (PAAC) \cite{DBLP:journals/corr/ClementeMC17}, the agents explore their own environments separately, and after a  number of samples are collected, the global parameters are updated synchronously. On the contrary, the coordinator will update the global model immediately after receiving the gradient from an arbitrary agent in asynchronous RL algorithms, rather than waiting for other agents. Several asynchronous RL algorithms are presented, including A3C \cite{10.5555/3045390.3045594}, Impala \cite{pmlr-v80-espeholt18a}, Ape-X \cite{DBLP:journals/corr/abs-1803-00933} and General Reinforcement Learning Architecture (Gorila) \cite{DBLP:journals/corr/NairSBAFMPSBPLM15}. From the perspective of technology development, HFRL can also be considered security-enhanced parallel RL. In parallel RL, multiple agents interact with a stochastic environment to seek the optimal policy for the same task \cite{DBLP:journals/corr/NairSBAFMPSBPLM15, 10.1007/978-3-540-77949-0_5}. By building a closed loop of data and knowledge in parallel systems, parallel RL helps determine the next course of action for each agent. The state and action representations are fed into a designed neural network to approximate the action value function \cite{8370297}. However, parallel RL typically transfers the experience of agent without considering privacy protection issues \cite{bookfederatedlearning}. In the implementation of HFRL, further restrictions accompany privacy protection and communication consumption to adapt to special scenarios, such as IoT applications \cite{9062302}. In addition, another point to consider is Non-IID data. In order to ensure convergence of the RL model, it is generally assumed in parallel RL that the states transitions in the environment follow the same distribution, \ie the environments of different agents are IID. But in actual scenarios, the situation faced by agents may differ slightly, so that the models of environments for different agents are not identically distributed. Therefore, HFRL needs to improve the generalization ability of the model compared with parallel RL to meet the challenges posed by Non-IID data.
	
	Based on the potential issues faced by the current RL technology, the advantages of HFRL can be summarized as follows.
	\begin{itemize}
		\item Enhancing training speed. In the case of a similar target task, multiple agents sharing training experiences gained from different environments can expedite the learning process. The local model rapidly evolves through aggregation and update algorithms to assess the unexplored environment. Moreover, the data obtained by different agents are independent, reducing correlations between the observed data. Furthermore, this also helps to solve the issue of unbalanced data caused by various restrictions.
		\item Improving the reliability of model. When the dimensions of the state of the environment are enormous or even uncountable, it is difficult for a single agent to train an optimal strategy for situations with extremely low occurrence probabilities. Horizontal agents are exploring independently while building a cooperative model to improve the local model's performance on rare states. 
		\item Mitigating the problems of devices heterogeneity. Different devices deploying RL agents in the HFRL architecture may have different computational and communication capabilities. Some devices may not meet the basic requirements for training, but strategies are needed to guide actions. HFRL makes it possible for all agents to obtain the shared model equally for the target task.
		\item Addressing the issue of non-identical environment. Considering the differences in the environment dynamics for the different agents, the assumption of IID data may be broken. Under the HFRL architecture, agents in not identically-distributed environment models can still cooperate to learn a federated model. In order to address the difference in environment dynamics, a personalized update algorithm of local model could be designed to minimize the impact of this issue.
		\item Increasing the flexibility of the system. The agent can decide when to participate in the cooperative system at any time, because HFRL allows asynchronous requests and aggregation of shared models. In the existing HFRL-based application, new agents also can apply for membership and benefit from downloading the shared model.
	\end{itemize}
	
	\subsection{Vertical federated reinforcement learning}
	
	In VFL, samples of multiple data sets have different feature spaces but these samples may belong to the same groups or common users. The training data of each participant are divided vertically according to their features. More general and accurate models can be generated by building heterogeneous feature spaces without releasing private information. VFRL applies  the methodology of VFL to RL and is suitable for POMDP scenarios where different RL agents are in the same environment but have different interactions with the environment. Specifically, different agents could have different observations that are only part of the global state. They could take actions from different action spaces and observe different rewards, or some agents even take no actions or cannot observe any rewards. Since the observation range of a single agent to the environment is limited, multiple agents cooperate to collect enough knowledge needed for decision making.  The role of FL in VFRL is to aggregate the partial features observed by various agents. Especially for those agents without rewards, the aggregation effect of FL greatly enhances the value of such agents in their interactions with the environment, and ultimately helps with the strategy optimization. It is worth noting that in VFRL the issue of privacy protection needs to be considered, \ie private data collected by some agents do not have to be shared with others. Instead, agents can transmit encrypted model parameters, gradients, or direct mid-product to each other. In short, the goal of VFRL is for agents  interacting with the same environment  to improve the performance of their policies and the effectiveness in learning them by sharing experiences without compromising the privacy.
	
	More formally, we denote $\left\{ \mathcal{F} _i \right\} _{i=1}^{N}$ as $N$ agents in VFRL, which interact with a global environment $\mathcal{E}$. The $i$-th agent $\mathcal{F} _i$ is located in the environment $\mathcal{E} _i=\mathcal{E}$, obtains the local partial observation $\mathcal{O} _i$, and can perform the set of actions $\mathcal{A} _i$. Different from HFRL, the state/observation and action spaces of two agents $\mathcal{F} _i$ and $\mathcal{F} _j$ may be not identical, but the aggregation of the state/observation spaces and action spaces of all the agents constitutes the global state and action spaces of the global environment $\mathcal{E}$. The conditions for VFRL can be defined as \ie
	
	$$
	\mathcal{O}_i\ne \mathcal{O}_j, \mathcal{A}_i\ne \mathcal{A}_j, \mathcal{E}_i=\mathcal{E}_j=\mathcal{E},
	\\
	\bigcup_{i=1}^N{\mathcal{O}_i=}\mathcal{S}, \bigcup_{i=1}^N{\mathcal{A}_i=}\mathcal{A},
	\\
	\forall i,j\in \left\{ \text{1,2,...,}N \right\} , i\ne j,
	$$
	
	\noindent where $\mathcal{S}$ and $\mathcal{A}$ denote the global state space and action space of all participant agents respectively. It can be seen that all the observations of the $N$ agents together constitute the global state space $\mathcal{S}$ of the environment $\mathcal{E}$. Besides, the environments $\mathcal{E}_i$ and $\mathcal{E}_j$ are the same environment $\mathcal{E}$. In most cases, there is a great difference between the observations of two agents $\mathcal{F} _i$ and $\mathcal{F} _j$.
	
\begin{figure}[htb]
		\centering
		\includegraphics[width=0.6\textwidth]{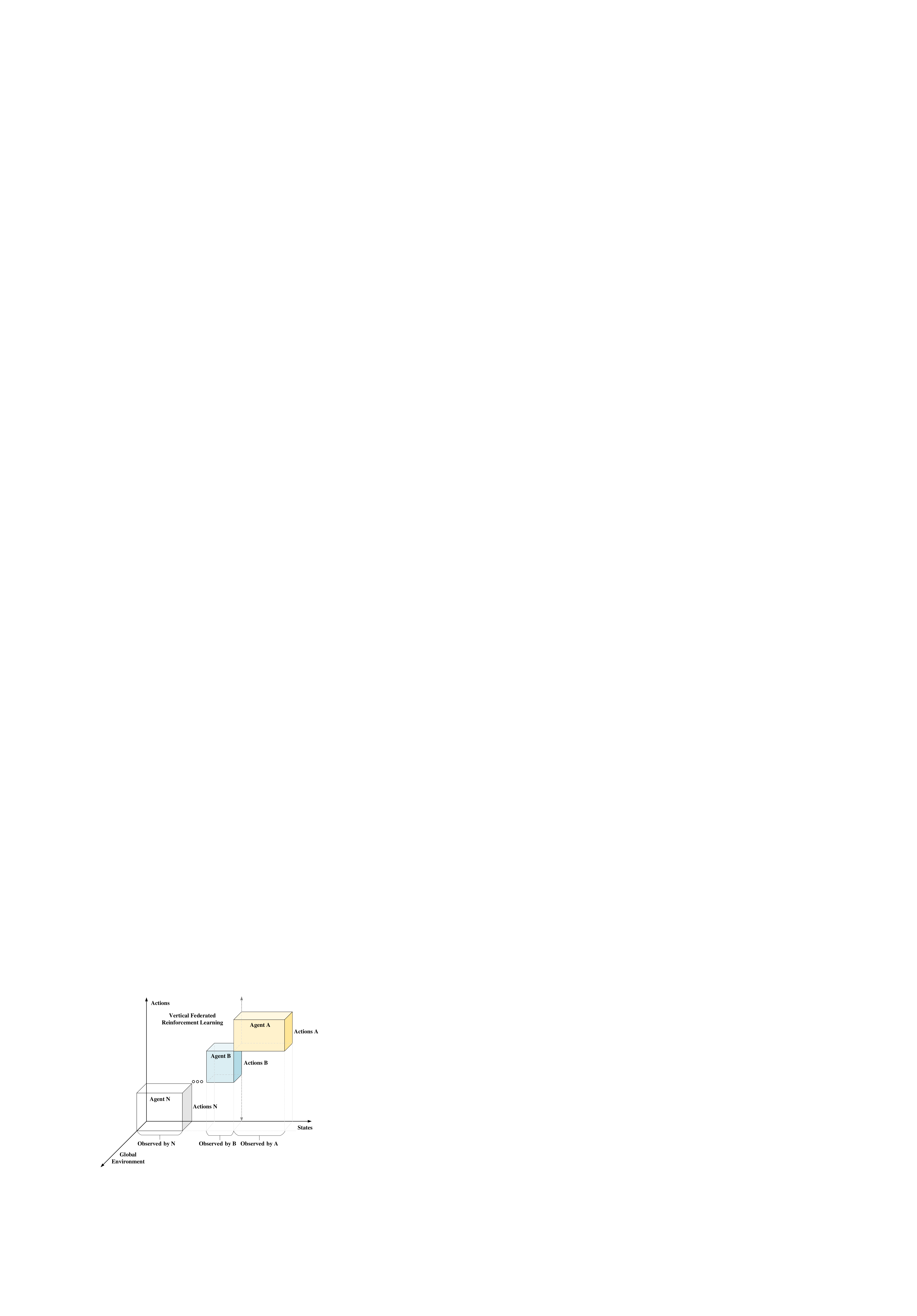}
		\caption{Illustration of vertical federated reinforcement learning (VFRL).}
		\label{fig-vfrl-ache}
	\end{figure}

	Figure \ref{fig-vfrl-ache} shows the architecture of VFRL.  The dataset and feature space in VFL are converted to the environment and state space respectively. VFL divides the dataset vertically according to the features of samples, and VFRL divides agents based on the state spaces observed from the global environment. Generally speaking, every agent has its local state which can be different from that of the other agents and the aggregation of these local partial states corresponds to the entire environment state \cite{zhuo2020federated}. In addition, after interacting with the environment, agents may generate their local actions which correspond to the labels in VFL. 
	
\begin{figure}[htb]
		\centering
		\includegraphics[width=0.8\textwidth]{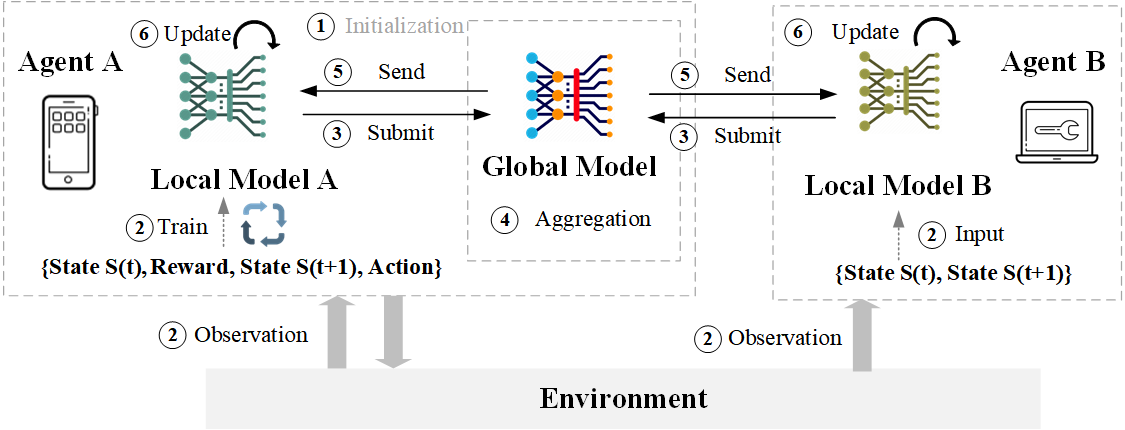}
		\caption{An example of vertical federated reinforcement learning (VFRL) architecture.}
		\label{fig-vfrl}
	\end{figure}

	Two types of agents can be defined for VFRL, \ie decision-oriented agents and support-oriented agents. Decision-oriented agents $\left\{ \mathcal{F} _i \right\} _{i=1}^{K}$ can interact with the environment $ \mathcal{E}$ based on their local state $\left\{ \mathcal{S} _i \right\} _{i=1}^{K}$ and action $\left\{ \mathcal{A} _i \right\} _{i=1}^{K}$. Meanwhile, support-oriented agents $\left\{ \mathcal{F} _i \right\} _{i=K+1}^{N}$ take no actions and receive no rewards but only the observations of the environment, \ie their local states $\left\{ \mathcal{S} _i \right\} _{i=K+1}^{N}$. In general, the following six steps, as shown in Figure \ref{fig-vfrl}, are the basic procedure for VFRL, \ie
	
	\begin{itemize}
		\item Step 1: Initialization is performed for all agent models. 
		\item Step 2: Agents obtain states from the environment. For decision-oriented agents, actions are obtained based on the local models, and feedbacks are obtained through interactions with the environment, \ie the states of the next time step and rewards. The data tuple of state-action-reward-state (SARS) is used to train the local models.
		\item Step 3: All agents calculate the mid-products of the local models and then transmit the encrypted mid-products to the federated model.
		\item Step 4: The federated model performs the aggregation calculation for mid-products and trains the federated model based on the aggregation results.
		\item Step 5: Federated model encrypts model parameters such as weight and gradient and passes them back to other agents.
		\item Step 6: All agents update their local models based on the received encrypted parameters.
	\end{itemize}
	
	As an example of VFRL, consider a microgrid (MG) system including household users, the power company, and the photovoltaic (PV) management company as the agents. All the agents observe the same MG environment while their local state spaces are quite different. The global states of the MG system generally consist of several dimensions/features, \ie State-of-Charge (SOC) of the batteries, load consumption of the household users, power generation from PV, etc. The household agents can obtain the SOC of their own batteries and their own load consumption, the power company can know the load consumption of all the users, and PV management company can know the power generation of PV. As to the action, the power company needs to make decisions on the power dispatch of the diesel generators (DG), and the household users can make decisions to manage their electrical utilities with demand response. Finally, the power company can observe rewards such as the cost of DG power generation, the balance between power generation and consumption, and the household users can observe rewards such as their electricity bill that is related to their power consumption. In order to learn the optimal policies, these agents need to communicate with each other to share their observations. However, PV managers do not want to expose their data to other companies, and household users also want to keep their consumption data private. In this way, VFRL is suitable to achieve this goal and can help improve policy decisions without exposing specific data. 
	
	Compared with HFRL, there are currently few works on VFRL. Zhuo \etal \cite{zhuo2020federated} present the federated deep reinforcement learning (FedRL) framework. The purpose of this paper is to solve the challenge where the feature space of states is small and the training data are limited. Transfer learning approaches in DRL are also solutions for this case. However, when considering the privacy-aware applications, directly transferring data or models should not be used. Hence, FedRL combines the advantage of FL with RL, which is suitable for the case when agents need to consider their privacy. FedRL framework assumes agents cannot share their partial observations of the environment and some agents are unable to receive rewards. It builds a shared value network, \ie MultiLayer Perceptron (MLP), and takes its own Q-network output and encryption value as input to calculate a global Q-network output. Based on the output of global Q-network, the shared value network and self Q-network are updated. Two agents are used in the FedRL algorithm, \ie agent $\alpha$ and $\beta$, which interact with the same environment. However, agent $\beta$ cannot build its own policies and rewards. Finally, FedRL is applied in two different games, \ie Grid-World and Text2Action, and achieves better results than the other baselines. Although the VFRL model in this paper only contains two agents, and the structure of the aggregated neural network model is relatively simple, we believe that it is a great attempt to first implement VFRL and verify its effectiveness.
	
	Multi-agent RL (MARL) is very closely related to VFRL. As the name implies, MARL takes into account the existence of multiple agents in the RL system. However, the empirical evaluation shows that applying the simple single-agent RL algorithms directly to scenarios of multiple agents cannot converge to the optimal solution, since the environment is no longer static from the perspective of each agent \cite{canese2021multi}. In specific, the action of each agent will affect the next state, thus affecting all agents in the future time step \cite{busoniu2008comprehensive}. Besides, the actions performed by one certain agent will yield different rewards depending on the actions taken by other agents. This means that agents in MARL correlate with each other, rather than being independent of each other. This challenge, called as the non-stationarity of the environment, is the main problem to be solved in the development of an efficient MARL algorithm \cite{zhang2021multi}. 
	
	MARL and VFRL both study the problem of multiple agents learning concurrently how to solve a task by interacting with the same environment \cite{stone2000multiagent}. Since MARL and VFRL have a large range of similarities, the review of MARL's related works is a very useful guide to help researchers summarize the research focus and better understand VFRL. There is abundant literature related to MARL. However, most MARL research \cite{szepesvari1999unified, littman2001value,tan1993multi,lauer2000algorithm} is based on a fully observed Markov Decision Process (MDP), where each agent is assumed to have the global observation of the system state \cite{zhang2021multi}. These MARL algorithms are not applicable to the case of POMDP where the observations of individual agents are often only a part of the overall environment \cite{monahan1982state}. Partial observability is a crucial consideration for the development of algorithms that can be applied to real-world problems \cite{oroojlooyjadid2019review}. Since VFRL is mainly oriented towards POMDP scenarios, it is more important to analyze the related works of MARL based on POMDP as the guidance of VFRL. 
	
	
	Agents in the above scenarios partially observe the system state and make decisions at each step to maximize the overall rewards for all agents, which can be formalized as a Decentralized Partially Observable Markov Decision Process (Dec-POMDP) \cite{bernstein2002complexity}. Optimally addressing a Dec-POMDP model is well known to be a very challenging problem. In the early works \cite{omidshafiei2017deep}, Omidshafiei \etal proposes a two-phase MT-MARL approach that concludes the methods of cautiously-optimistic learners for action-value approximation and Concurrent Experience Replay Trajectories (CERTs) as the experience replay targeting sample-efficient and stable MARL. The authors also apply the recursive neural network (RNN) to estimate the non-observed state and hysteretic Q-learning to address the problem of non-stationarity in Dec-POMDP. Han \etal \cite{han2019ipomdp} designs a neural network architecture, IPOMDP-net, which extends QMDP-net planning algorithm \cite{karkus2017qmdpnet} to MARL settings under POMDP. Besides, Mao \etal \cite{9303801} introduces the concept of information state embedding to compress agents’ histories and proposes an RNN model combining the state embedding. Their method, \ie embed-then-learn pipeline, is universal since the embedding can be fed into any existing partially observable MARL algorithm as the black-box. In the study from Mao \etal \cite{mao2018modelling}, the proposed Attention MADDPG (ATT-MADDPG) has several critic networks for various agents under POMDP. A centralized critic is adopted to collect the observations and actions of the teammate agents. Specifically, the attention mechanism is applied to enhance the centralized critic. The final introduced work is from Lee \etal \cite{lee2021multi}. They present an augmenting MARL algorithm based on pretraining to address the challenge in disaster response. It is interesting that they use behavioral cloning (BC), a supervised learning method where agents learn their policy from demonstration samples, as the approach to pretrain the neural network. BC can generate a feasible Dec-POMDP policy from demonstration samples, which offers advantages over plain MARL in terms of solution quality and computation time.

	Some MARL algorithms also concentrate on the communication issue of POMDP. In the study from Sukhbaatar \cite{sukhbaatar2016learning}, communication between the agents is performed for a number of rounds before their action is selected. The communication protocol is learned concurrently with the optimal policy. Foerster \etal \cite{foerster2016learning} proposes a deep recursive network architecture, \ie Deep Distributed Recurrent Q-Network (DDRQN), to address the communication problem in a multi-agent partially-observable setting. This work makes three fundamental modifications to previous algorithms. The first one is last-action inputs, which let each agent access its previous action as an input for the next time-step. Besides, inter-agent weight sharing allows diverse behavior between agents, as the agents receive different observations and thus evolve in different hidden states. The final one is disabling experience replay, which is because the non-stationarity of the environment renders old experiences obsolete or misleading. Foerster \etal \cite{foerster2016learning} considers the communication task of fully cooperative, partially observable, sequential multi-agent decision-making problems. In their system model, each agent can receive a private observation and take actions that affect the environment. In addition, the agent can also communicate with its fellow agents via a discrete limited-bandwidth channel. Despite the partial observability and limited channel capacity, authors achieved the task that the two agents could discover a communication protocol that enables them to coordinate their behavior based on the approach of deep recurrent Q-networks.
	
	While there are some similarities between MARL and VFRL, several important differences have to be paid attention to, \ie
	
	\begin{itemize}
		\item VFRL and some MARL algorithms are able to address similar problems, \eg the issues of POMDP. However, there are differences between the solution ideas between two algorithms. Since VFRL is the product of applying VFL to RL, the FL component of VFRL focuses more on the aggregation of partial features, including states and rewards, observed by different agents since VFRL inception. Security is also an essential issue in VFRL. On the contrary, MARL may arise as the most natural way of adding more than one agent in a RL system \cite{bucsoniu2010multi}. In MARL, agents not only interact with the environment, but also have complex interactive relationships with other agents, which creates a great obstacle to the solution of policy optimization. Therefore, the original intentions of two algorithms are different.
		
		\item Two algorithms are slightly different in terms of the structure. The agents in MARL must surely have the reward even some of them may not have their own local actions. However, in some cases, the agents in VFRL are not able to generate a corresponding operation policy, so in these cases, some agents have no actions and rewards \cite{zhuo2020federated}. Therefore, VFRL can solve more extensive problems that MARL is not capable of solving. 
		
		\item Both two algorithms involve the communication problem between agents. In MARL, information such as the states of other agents and model parameters can be directly and freely propagated among agents. During communication, some MARL methods such as DDRQN in the work of Foerster \etal \cite{foerster2016learning} consider the previous action as an input for the next time-step state. Weight sharing is also allowed between agents. However, VFRL assumes states cannot be shared among agents. Since these agents do not exchange experience and data directly, VFRL focuses more on security and privacy issues of communication between agents, as well as how to process mid-products transferred by other agents and aggregate federated models. 
	\end{itemize}
	
	In summary, as a potential and notable algorithm, VFRL has several advantages as follows, \ie
	\begin{itemize}
		\item Excellent privacy protection. VFRL inherits the FL algorithm's idea of data privacy protection, so for the task of multiple agents cooperation in the same environment, information interaction can be carried out confidently to enhance the learning efficiency of RL model. In this process, each participant does not have to worry about any leakage of raw real-time data. 
		
		\item Wide application scenarios. With appropriate knowledge extraction methods, including algorithm design and system modeling, VFRL can solve more real-world problems compared with MARL algorithms. This is because VFRL can consider some agents that cannot generate rewards into the system model, so as to integrate their partial observation information of the environment based on FL while protecting privacy, train a more robust RL agent, and further improve learning efficiency. 
	\end{itemize}
	
	\subsection{Other types of FRL}
	
	The above HFRL or VFRL algorithms borrow ideas from FL for federation between RL agents. Meanwhile, there are also some existing works on FRL that are less affected by FL. Hence, they do not belong to either HFRL or VFRL, but federation between agents is also implemented.
	
	The study from Hu \etal \cite{9408573} is a typical example, which proposes a reward shaping based general FRL algorithm, called Federated Reward Shaping (FRS). It uses reward shaping to share federated information to improve policy quality and training speed. FRS adopts the server-client architecture. The server includes the federated model, while each client completes its own tasks based on the local model. This algorithm can be combined with different kinds of RL algorithms. However, it should be noted that FRS focuses on reward shaping, this algorithm cannot be used when there is no reward in some agents in VFRL. In addition, FRS performs knowledge aggregation by sharing high-level information such as reward shaping value or embedding between client and server instead of sharing experience or policy directly. The convergence of FRS is also guaranteed since only minor changes are made during the learning process, which is the modification of the reward in the replay buffer.
	
	As another example, Anwar \etal \cite{anwar2021multi} achieves federation between agents by smoothing the average weight. This work analyzes the Multi-task FRL algorithms (MT-FedRL) with adversaries. Agents only interact and make observations in their environment, which can be featured by different MDPs. Different from HFRL, the state and action spaces do not need to be the same in these environments. The goal of MT-FedRL is to learn a unified policy, which is jointly optimized across all of the environments. MT-FedRL adopts policy gradient methods for RL. In other words, policy parameter is needed to learn the optimal policy. The server-client architecture is also applied and all agents should share their own information with a centralized server. The role of non-negative smoothing average weights is to achieve a consensus among the agents’ parameters. As a result, they can help to incorporate the knowledge from other agents as the process of federation.
	
	\section{Applications of FRL}
	In this section, we provide an extensive discussion of the application of FRL in a variety of tasks, such as edge computing, communications, control optimization, attack detection, \etc. This section is aimed at enabling  readers to understand the applicable scenarios and research status of FRL.
	
	\subsection{FRL for edge computing}
	In recent years, edge equipment, such as BSs and Road Side Units (RSUs), has been equipped with increasingly advanced communication, computing  and storage capabilities. As a result, edge computing is proposed to delegating more tasks to edge equipment in order to reduce the communication load and reduce the corresponding delay. However, the issue of privacy protection remains challenging since it may be untrustworthy for the data owner to hand off their private information to a third-party edge server \cite{9060868}. FRL offers a potential solution for achieving privacy-protected intelligent edge computing, especially in decision-making tasks like caching and offloading. Additionally, the multi-layer processing architecture of edge computing is also suitable for implementing FRL through the server-client model. Therefore, many researchers have focused on applying FRL to edge computing.
	
	The distributed data of large-scale edge computing architecture makes it possible for FRL to provide distributed intelligent solutions to achieve resource optimization at the edge. For mobile edge networks, a potential FRL framework is presented for edge system \cite{8770530}, named as “In-Edge AI”, to address optimization of mobile edge computing, caching, and communication problems. The authors also propose some ideas and paradigms for solving these problems by using DRL and Distributed DRL. To carry out dynamic system-level optimization and reduce the unnecessary transmission load, “In-Edge AI” framework takes advantage of the collaboration among edge nodes to exchange learning parameters for better training and inference of models. It has been evaluated that the framework has high performance and relatively low learning overhead, while the mobile communication system is cognitive and adaptive to the environment. The paper provides good prospects for the application of FRL to edge computing, but there are still many challenges to overcome, including the adaptive improvement of the algorithm, and the training time of the model from scratch \etc.
	
	Edge caching has been considered a promising technique for edge computing to meet the growing demands for next-generation mobile networks and beyond. Addressing the adaptability and collaboration challenges of the dynamic network environment, Wang \etal \cite{9252973} proposes a Device-to-device (D2D)-assisted heterogeneous collaborative edge caching framework. User equipment (UE) in a mobile network uses the local DQN model to make node selection and cache replacement decisions based on network status and historical information. In other words, UE decides where to fetch content and which content should be replaced in its cache list. The BS calculates aggregation weights based on the training evaluation indicators from UE. To solve the long-term mixed-integer linear programming problem, the attention-weighted federated deep reinforcement learning (AWFDRL) is presented, which optimizes the aggregation weights to avoid the imbalance of the local model quality and improve the learning efficiency of the DQN. The convergence of the proposed algorithm is verified and simulation results show that the AWFDRL framework can perform well on average delay, hit rate, and offload traffic.
	
	A federated solution for cooperative edge caching management in fog radio access networks (F-RANs) is proposed \cite{9473609}. Both edge computing and fog computing involve bringing intelligence and processing to the origins of data. The key difference between the two architectures is where the computing node is positioned. A dueling deep Q-network based cooperative edge caching method is proposed to overcome the dimensionality curse of RL problem and improve caching performance. Agents are developed in fog access points (F-APs) and allowed to build a local caching model for optimal caching decisions based on the user content request and the popularity of content. HFRL is applied to aggregate the local models into a global model in the cloud server. The proposed method outperforms three classical content caching methods and two RL algorithms in terms of reducing content request delays and increasing cache hit rates. 
	
	For edge-enabled IoT, Majidi \etal \cite{9447004} proposes a dynamic cooperative caching method based on hierarchical federated deep reinforcement learning (HFDRL), which is used to determine which content should be cached or evicted by predicting future user requests. Edge devices that have a strong relationship are grouped into a cluster and one head is selected for this cluster. The BS trains the Q-value based local model by using BS states, content states, and request states. The head has enough processing and caching capabilities to deal with model aggregation in the cluster. By categorizing edge devices hierarchically, HFDRL improves the response time delay to keeps both small and large clusters from experiencing the disadvantages they could encounter. Storage partitioning allows content to be stored in clusters at different levels using the storage space of each device. The simulation results show the proposed method using MovieLens datasets improves the average content access delay and the hit rate.
	
	Considering the low latency requirements and privacy protection issue of IoV, the study of efficient and secure caching methods has attracted many researchers. An FRL-empowered task caching problem with IoV has been analyzed by Zhao \etal \cite{9500714}. The work proposes a novel cooperative caching algorithm (CoCaRL) for vehicular networks with multi-level FRL to dynamically determine which contents should be replaced and where the content requests should be served. This paper develops a two-level aggregation mechanism for federated learning to speed up the convergence rate and reduces communication overhead, while DRL task is employed to optimize the cooperative caching policy between RSUs of vehicular networks. Simulation results show that the proposed algorithm can achieve a high hit rate, good adaptability and fast convergence in a complex environment.
	
	Apart from caching services, FRL has demonstrated its strong ability to facilitate resource allocation in edge computing. In the study from Zhu \etal \cite{9426913}, the authors specifically focus on the data offloading task for Mobile Edge Computing (MEC) systems. To achieve joint collaboration, the heterogeneous multi-agent actor-critic (H-MAAC) framework is proposed, in which edge devices independently learn the interactive strategies through their own observations. The problem is formulated as a multi-agent MDP for modeling edge devices' data allocation strategies, \ie moving the data, locally executing or offloading to a cloud server. The corresponding joint cooperation algorithm that combines the edge federated model with the multi-agent actor-critic RL is also presented. Dual lightweight neural networks are built, comprising original actor/critic networks and target actor/critic networks.
	
	Blockchain technology has also attracted lot attention from researchers in edge computing fields since it is able to provide reliable data management within the massive distributed edge nodes. In the study from Yu \etal \cite{yu_when_2020}, the intelligent ultra-dense edge computing (I-UDEC) framework is proposed, integrating with blockchain and RL technologies into 5G ultra-dense edge computing networks. In order to achieve low overhead computation offloading decisions and resource allocation strategies, authors design a  two-timescale deep reinforcement learning (2Ts-DRL) approach, which consists of a fast-timescale and a slow-timescale learning process. The target model can be trained in a distributed manner via FL architecture, protecting the privacy of edge devices.
	
	Additionally, to deal with the different types of optimization tasks, variants of FRL are being studied. Zhu \etal presents a resource allocation method for edge computing systems, called concurrent federated reinforcement learning (CFRL) \cite{9454444}. The edge node continuously receives tasks from serviced IoT devices and stores those tasks in a queue. Depending on its own resource allocation status, the node determines the scheduling strategy so that all tasks are completed as soon as possible. In case the edge host does not have enough available resources for the task, the task can be offloaded to the server.  Contrary to the definition of the central server in the basic FRL, the aim of central server in CFRL is to complete the tasks that the edge nodes cannot handle instead of aggregating local models. Therefore, the server needs to train a special resource allocation model based on its own resource status, forwarded tasks and unique rewards. The main idea of CFRL is that edge nodes and the server cooperatively participate in all task processing in order to reduce total computing time and provide a degree of privacy protection.
	
	\subsection{FRL for communication networks}
	In parallel with the continuous evolution of communication technology, a number of heterogeneous communication systems are also being developed to adapt to different scenarios. Many researchers are also working toward intelligent management of communication systems. The traditional ML-based management methods are often inefficient due to their centralized data processing architecture and the risk of privacy leakage \cite{9415623}. FRL can play an important role in services slicing and access controlling to replace centralized ML methods.
	
	In communication network services, Network Function Virtualization (NFV) is a critical component of achieving scalability and flexibility. Huang \etal proposes a novel scalable service function chains orchestration (SSCO) scheme for NFV-enabled networks via FRL\cite{9468364}. In the work, a federated-learning-based framework for training global learning, along with a time-variant local model exploration, is designed for scalable SFC orchestration. It prevents data sharing among stakeholders and enables quick convergence of the global model. To reduce communication costs, SSCO allows the parameters of local models to be updated just at the beginning and end of each episode through distributed clients and the cloud server. A DRL approach is used to map Virtual Network Functions (VNFs) into networks with local knowledge of resources and instantiation cost. In addition, the authors also propose a loss-weight-based mechanism for generation and exploitation of reference samples for training in replay buffers, avoiding the strong relevance of each sample. Simulation results demonstrate that SSCO can significantly reduce placement errors and improve resource utilization ratios to place time-variant VNFs, as well as achieving desirable scalability.
	
	Network slicing (NS) is also a form of virtual network architecture to support divergent requirements sustainably. The work from Liu \etal \cite{9237167} proposes a device association scheme (such as access control and handover management) for Radio Access Network (RAN) slicing by exploiting a hybrid federated deep reinforcement learning (HDRL) framework. In view of the large state-action space and variety of services, HDRL is designed with two layers of model aggregations. Horizontal aggregation deployed on BSs is used for the same type of service. Generally, data samples collected by different devices within the same service have similar features. The discrete-action DRL algorithm, \ie DDQN, is employed to train the local model on individual smart devices. BS is able to aggregate model parameters and establish a cooperative global model. Vertical aggregation developed on the third encrypted party is responsible for the services of different types.  In order to promote collaboration between devices with different tasks, authors aggregate local access features to form a global access feature, in which the data from different flows is strongly correlated since different data flows are competing for radio resources with each other. Furthermore, the Shapley value \cite{9006179}, which represents the average marginal contribution of a specific feature across all possible feature combinations, is used to reduce communication cost in vertical aggregation based on the global access feature. Simulation results show that HDRL can improve network throughput and communication efficiency.
	
	The Open Radio Access Network (O-RAN) has emerged as a paradigm for supporting multi-class wireless services in 5G and beyond networks. To deal with the two critical issues of load balance and handover control, Cao \etal proposes a federated DRL-based scheme to train the model for user access control in the O-RAN \cite{9500603}. Due to the mobility of UEs and the high cost of the handover between BSs, it is necessary for each UE to access the appropriate BS to optimize its throughput performance. As independent agents, UEs make access decisions with assistance from a global model server, which updates global DQN parameters by averaging DQN parameters of selected UEs. Further, the scheme proposes only partially exchanging DQN parameters to reduce communication overheads, and using the dueling structure to allow convergence for independent agents. Simulation results demonstrate that the scheme increases long-term throughput while avoiding frequent handovers of users with limited signaling overheads.
	
	The issue of optimizing user access is important in wireless communication systems. FRL can provide interesting solutions for enabling efficient and privacy-enhanced management of access control. Zhang \etal \cite{9348485} studies the problem of multi-user access in WIFI networks. In order to mitigate collision events on channel access, an enhanced multiple access mechanism based on FRL is proposed for user-dense scenarios. In particular, distributed stations train their local q-learning networks through channel state, access history and feedback from central Access Point (AP). AP uses the central aggregation algorithm to update the global model every period of time and broadcast it to all stations. In addition, a Monte Carlo (MC) reward estimation method for the training phase of local model is introduced, which allocates more weight to the reward of that current state by reducing the previous cumulative reward. 
	
	FRL is also studied for Intelligent Cyber-Physical Systems (ICPS), which aims to meet the requirements of intelligent applications for high-precision, low-latency analysis of big data. In light of the heterogeneity brought by multiple agents, the central RL-based resource allocation scheme has non-stationary issues and does not consider privacy issues. Therefore, the work from Xu \etal \cite{9434397} proposes a multi-agent FRL (MA-FRL) mechanism which synthesizes a good inferential global policy from encrypted local policies of agents without revealing private information. The data resource allocation and secure communication problems are formulated as a Stackelberg game with multiple participants, including near devices (NDs), far devices (FDs) and relay devices (RDs). Take into account the limited scope of the heterogeneous devices, the authors model this multi-agent system as a POMDP. Furthermore, it is proved that MA-FRL is $\mu$-strongly convex and $\beta$-smooth and derives its convergence speed in expectation.
	
	Zhang \etal \cite{8944302} pays attention to the challenges in cellular vehicle-to-everything (V2X) communication for future vehicular applications. A joint optimization problem of selecting the transmission mode and allocating the resources is presented. This paper proposes a decentralized DRL algorithm for maximizing the amount of available vehicle-to-infrastructure capacity while meeting the latency and reliability requirements of vehicle-to-vehicle (V2V) pairs. Considering limited local training data at vehicles, the federated learning algorithm is conducted on a small timescale. On the other hand, the graph theory-based vehicle clustering algorithm is conducted on a large timescale.
	
	The development of communication technologies in extreme environments is important, including deep underwater exploration. The architecture and philosophy of FRL are applied to smart ocean applications in the study of Kwon \cite{9067847}. To deal with the nonstationary environment and unreliable channels of underwater wireless networks, the authors propose a  multi-agent DRL-based algorithm that can realize FL computation with Internet-of-Underwater-Things (IoUT) devices in the ocean environment. The cooperative model is trained by MADDPG for cell association and resource allocation problems. As for downlink throughput, it is found that the proposed MADDPG-based algorithm performed 80\% and 41\% better than the standard actor-critic and DDPG algorithms, respectively.
	
	\subsection{FRL for control optimization}
	Reinforcement learning based control schemes are considered as one of the most effective ways to learn a nonlinear control strategy in complex scenarios, such as robotics. Individual agent's exploration of the environment is limited by its own field of vision and usually needs a great deal of training to obtain the optimal strategy. The FRL-based approach has emerged as an appealing way to realize control optimization without exposing agent data or compromising privacy.
	
	Automated control of robots is a typical example of control optimization problems. Liu \etal \cite{DBLP:journals/corr/abs-1901-06455} discusses robot navigation scenarios and focuses on how to make robots transfer their experience so that they can make use of prior knowledge and quickly adapt to changing environments. As a solution, a cooperative learning architecture, called LFRL, is proposed for navigation in cloud robotic systems. Under the FRL-based architecture, the authors propose a corresponding knowledge fusion algorithm to upgrade the shared model deployed on the cloud. In addition, the paper also discusses the problems and feasibility of applying transfer learning algorithms to different tasks and network structures between the shared model and the local model.
	
	FRL is combined with autonomous driving of robotic vehicles in the study of Liang \etal \cite{liang_federated_2019}. To reach rapid training from a simulation environment to a real-world environment, Liang \etal presents a Federated Transfer Reinforcement Learning (FTRL) framework for knowledge extraction where all the vehicles make corresponding actions with the knowledge learned by others. The framework can potentially be used to train more powerful tasks by pooling the resources of multiple entities without revealing raw data information in real-life scenarios. To evaluate the feasibility of the proposed framework, authors perform  real-life experiments on steering control tasks for collision avoidance of autonomous driving robotic cars and it is demonstrated that the framework has superior performance to the non-federated local training process. Note that the framework can be considered an extension of HFRL, because the target tasks to be accomplished are highly-relative and all observation data are pre-aligned.
	
	FRL also appears as an attractive approach for enabling intelligent control of IoT devices without revealing private information. Lim \etal proposes a FRL architecture which allows agents working on independent IoT devices to share their learning experiences with each other, and transfer the policy model parameters to other agents \cite{lim_federated_2020}. The aim is to effectively control multiple IoT devices of the same type but with slightly different dynamics. Whenever an agent meets the predefined criteria, its mature model will be shared by the server with all other agents in training. The agents continue training based on the shared model until the local model converges in the respective environment. The actor-critical proximal policy optimization (Actor-Critic PPO) algorithm is integrated into the control of multiple rotary inverted pendulum (RIP) devices. The results show that the proposed architecture facilitates the learning process and if more agents participate the learning speed can be improved. In addition, Lim \etal \cite{9439484} uses FRL architecture based on a multi-agent environment to solve the problems and limitations of RL for applications to the real-world problems. The proposed federation policy allows multiple agents to share their learning experiences to get better learning efficacy. The proposed scheme adopts Actor-Critic PPO algorithm for four types of RL simulation environments from OpenAI Gym as well as RIP in real control systems. Compared to a previous real-environment study, the scheme enhances learning performance by approximately 1.2 times.
	
	\subsection{FRL for attack detection}
	With the heterogeneity of services and the sophistication of threats, it is challenging to detect these attacks using traditional methods or centralized ML-based methods, which have a high false alarm rate and do not take privacy into account. FRL offers a powerful alternative to detecting attacks and provides support for network defense in different scenarios.
	
	Because of various constraints, IoT applications have become a primary target for malicious adversaries that can disrupt normal operations or steal confidential information. In order to address the security issues in flying ad-hoc network (FANET), Mowla \etal proposes an adaptive FRL-based jamming attack defense strategy for unmanned aerial vehicles (UAVs) \cite{9143577}. A model-free Q-learning mechanism is developed and deployed on distributed UAVs to cooperatively learn detection models for jamming attacks. According to the results, the average accuracy of the federated jamming detection mechanism, employed in the proposed defense strategy, is 39.9\% higher than the distributed mechanism when verified with the CRAWDAD standard and the ns-3 simulated FANET jamming attack dataset.
	
	An efficient traffic monitoring framework, known as DeepMonitor, is presented in the study of Nguyen \etal \cite{9508440} to provide fine-grained traffic analysis capability at the edge of Software Defined Network (SDN) based IoT networks. The agents deployed in edge nodes consider the different granularity-level requirements and their maximum flow-table capacity to achieve the optimal flow rule match-field strategy. The control optimization problem is formulated as the MDP and a federated DDQN algorithm is developed to improve the learning performance of agents. The results show that the proposed monitoring framework can produce reliable traffic granularity at all levels of traffic granularity and substantially mitigate the issue of flow-table overflows. In addition, the Distributed Denial Of Service (DDoS) attack detection performance of an intrusion detection system can be enhanced by up to 22.83\% by using DeepMonitor instead of FlowStat. 
	
	In order to reduce manufacturing costs and improve production efficiency, the Industrial Internet of Things (IIoT) is proposed as a potentially promising  research direction. It is a challenge to implement anomaly detection mechanisms in IIoT applications with data privacy protection. Wang \etal proposes a reliable anomaly detection strategy for IIoT using FRL techniques\cite{9409113}. In the system framework, there are four entities involved in establishing the detection model, \ie the Global Anomaly Detection Center (GADC), the Local Anomaly Detection Center (LADC), the Regional Anomaly Detection Center (RADC), and the users. The anomaly detection is suggested to be implemented in two phases, including anomaly detection for RADC and users. Especially, the GADC can build global RADC anomaly detection models based on local models trained by LADCs. Different from RADC anomaly detection based on action deviations, user anomaly detection is mainly concerned with privacy leakage and is employed by RADC and GADC. Note that the DDPG algorithm is applied for local anomaly detection model training.
	
	\subsection{FRL for other applications}
	Due to the outstanding performance of training efficiency and privacy protection, many researchers are exploring the possible applications of FRL.
	
	FL has been applied to realize distributed energy management in IoT applications. In the revolution of smart home, smart meters are deployed in the advanced metering infrastructure (AMI) to monitor and analyze the energy consumption of users in real-time. As an example \cite{9247266}, the FRL-based approach is proposed for the energy management of multiple smart homes with solar PVs, home appliances, and energy storage. Multiple local home energy management systems (LHEMSs) and a global server (GS) make up FRL architecture of the smart home. DRL agents for LHEMSs construct and upload local models to the GS by using energy consumption data. The GS updates the global model based on local models of LHEMSs using the federated stochastic gradient descent (FedSGD) algorithm. Under heterogeneous home environments, simulation results indicate that the proposed approach outperforms others when it comes to convergence speed, appliance energy consumption, and the number of agents.
	
	Moreover, FRL offers an alternative to share information with low latency and privacy preservation. The collaborative perception of vehicles provided by IoV can greatly enhance the ability to sense things beyond their line of sight, which is important for autonomous driving. Region quadtrees have been proposed as a storage and communication resource-saving solution for sharing perception information \cite{10.1145/356924.356930}. It is challenging to tailor the number and resolution of transmitted quadtree blocks to bandwidth availability. In the framework of FRL, Mohamed \etal presents a quadtree-based point cloud compression mechanism to select cooperative perception messages\cite{9443539}. Specifically, over a period of time, each vehicle covered by an RSU transfers its latest network weights with the RSU, which then averages all of the received model parameters and broadcasts the result back to the vehicles. Optimal sensory information transmission (\ie quadtree blocks) and appropriate resolution levels for a given vehicle pair are the main objectives of a vehicle. The dueling and branching concepts are also applied to overcome the vast action space inherent in the formulation of the RL problem. Simulation results show that the learned policies achieve higher vehicular satisfaction and the training process is enhanced by FRL.

	\subsection{Lessons Learned}
	In the following, we summarize the major lessons learned from this survey in order to provide a comprehensive understanding of current research on FRL applications.
	
	\subsubsection{Lessons learned from the aggregation algorithms}
	
	The existing FRL literature usually uses classical DRL algorithms, such as DQN and DDPG, at the participants, while the gradients or parameters of the critic and/or actor networks are periodically reported synchronously or asynchronously by the participants to the coordinator. The coordinator then aggregates the parameters or gradients and sends the updated values to the participants. In order to meet the challenges presented by different scenarios, the aggregation algorithms have been designed as a key feature of FRL. In the original FedAvg algorithm \cite{DBLP:journals/corr/McMahanMRA16}, the number of samples in a participant's dataset determines its influence on the global model. In accordance with this idea, several papers propose different methods to calculate the weights in the aggregation algorithms according to the requirement of application. In the study from Lim \etal \cite{9439484}, the aggregation weight is derived from the average of the cumulative rewards of the last ten episodes. Greater weights are placed on the models of those participants with higher rewards. In contrast to the positive correlation of reward, Huang \etal \cite{9468364} takes the error rate of action as an essential factor to assign weights for participating in the global model training. In D2D -assisted edge caching, Wang \etal \cite{9252973} uses the reward and some device-related indicators as the measurement to evaluate the local model’s contribution to the global model. Moreover, the existing FRL methods based on offline DRL algorithms, such DQN and DDPG, usually use experience replay. Sampling random batch from replay memory can break correlations of continuous transition tuples and accelerate the training process. To arrive at an accurate evaluation of the participants, the paper \cite{8944302} calculates the aggregation weight based on the size of the training batch in each iteration.
	
	The above aggregation methods can effectively deal with the issue of data imbalance and performance discrepancy between participants, but it is hard for participants to cope with subtle environmental differences. According to the paper \cite{lim_federated_2020}, as soon as a participant reaches the predefined criteria in its own environment, it should stop learning and send its model parameters as a reference to the remaining individuals. Exchanging mature network models (satisfying terminal conditions) can help other participants complete their training quickly. Participants in other similar environments can continue to use FRL for further updating their parameters to achieve the desired model performance according to their individual environments. Liu \etal \cite{DBLP:journals/corr/abs-1901-06455} also suggests that the sharing global model in the cloud is not the final policy model for local participants. An effective transfer learning should be applied to resolve the structural difference between the shared network and private network.

	\subsubsection{Lessons learned from the relationship between FL and RL}
	
	In most of the literature on FRL, FL is used to improve the performance of RL. With FL, the learning experience can be shared among decentralized multiple parties while ensuring privacy and scalability without requiring direct data offloading to servers or third parties. Therefore, FL can expand the scope and enhance the security of RL. Among the applications of FRL, most researchers focus on the communication network system due to its robust security requirements, advanced distributed architecture, and a variety of decision-making tasks. Data offloading \cite{9426913} and caching \cite{9252973} solutions powered by distributed AI are available from FRL. In addition, with the ability to detect a wide range of attacks and support defense solutions, FRL has emerged as a strong alternative for performing distributed learning for security-sensitive scenarios. Enabled by the privacy-enhancing and cooperative features, detection and defense solutions can be learned quickly where multiple participants join to build a federated model \cite{9143577,9409113}. FRL can also provide viable solutions to realize intelligence for control systems with many applied domains such as robotics \cite{DBLP:journals/corr/abs-1901-06455} and autonomous driving \cite{liang_federated_2019} without data exchange and privacy leakage. The data owners (robot or vehicle) may not trust the third-party server and therefore hesitate to upload their private information to potentially insecure learning systems. Each participant of FRL runs a separate RL model for determining its own control policy and gains experience by sharing model parameters, gradients or losses.
	
	
	Meanwhile, RL may have the potential to optimize FL schemes and improve the efficiency of training. Due to the unstable  network connectivity, it is not practical for FL to update and aggregate models simultaneously across all participants. Therefore, Wang \etal \cite{wang_optimizing_2020} proposes a RL-based control framework that intelligently chooses the participants to participate in each round of FL with the aim to speed up convergence. Similarly, Zhang \etal \cite{9452166} applies RL to pre-select a set of candidate edge participants, and then determine reliable edge participants through social attribute perception. In IoT or IoV scenarios, due to the heterogeneous nature of participating devices, different computing and communication resources are available to them. RL can speed up training by coordinating the allocation of resources between participants. Zhan \etal \cite{9384231} defines the L4L (Learning for Learning) concept, \ie use RL to improve FL. Using the heterogeneity of participants and dynamic network connections, this paper investigates a computational resource control problem for FL that simultaneously considers learning time and energy efficiency. An experience-driven resource control approach based on RL is presented to derive the near-optimal strategy with only the participants' bandwidth information in the previous training rounds. In addition, as with any other ML algorithm, FL algorithms are vulnerable to malicious attacks. RL has been studied to defend against attacks in various scenarios, and it can also enhance the security of FL. The paper \cite{9469426} proposes a reputation-aware RL (RA-RL) based selection method to ensure that FL is not disrupted. The participating devices' attributes, including computing resources and trust values, \etc, are used as part of the environment in RL. In the aggregation of the global model, devices with high reputation levels will have a greater chance of being considered to reduce the effects of malicious devices mixed into FL.
	
	\subsubsection{Lessons learned from categories of FRL}
	
	As discussed above, FRL can be divided into two main categories, \ie HFRL and VFRL. Currently, most of the existing research is focused on HFRL, while little attention is devoted to VFRL. The reason for this is that HFRL has obvious application scenarios, where multiple participants have similar decision-making tasks with individual environments, such as caching allocation \cite{9062302}, offloading optimization \cite{8728285}, and attack monitoring \cite{9508440}. The participants and coordinator only need to train a similar model with the same state and action spaces. Consequently, the algorithm design can be implemented and the training convergence can be verified relatively easily. On the other hand, even though VFRL has a higher degree of technical difficulty at the algorithm design level, it also has a wide range of possible applications. In a multi-agent scenario, for example, a single agent is limited by its ability to observe only part of the environment, whereas the transition of the environment is determined by the behavior of all the agents. Zhuo \etal \cite{zhuo2020federated} assumes agents cannot share their partial observations of the environment and some agents are unable to receive rewards. The federated Q-network aggregation algorithm between two agents is proposed for VFRL. The paper \cite{9237167} specifically applies both HFRL and VFRL for radio access network slicing.  For the same type of services, similar data samples are trained locally at participating devices, and BSs perform horizontal aggregation to integrate a cooperative access model by adopting an iterative approach. The terminal device also can optimize the selection of base stations and network slices based on the global model of VFRL, which aggregates access features generated by different types of services on the third encrypted party. The method improves the device's ability to select the appropriate access points when initiating different types of service requests under restrictions regarding privacy protection. The feasible implementation of VFRL also provides guidance for future research.

	\section{Open issues and Future Research Directions}
	As we presented in the previous section, FRL serves an increasingly important role as an enabler of various applications. While the FRL-based approach possesses many advantages, there are a number of critical open issues to consider for future implementation. Therefore, this section focuses on several key challenges, including those inherited from FL such as security and communication issues, as well as those unique to FRL. Research on tackling these issues offers interesting directions for the future.
	
	\subsection{Learning convergence in HFRL}
	In realistic HFRL scenarios, while the agents perform similar tasks, the inherent dynamics for the different environments in which the agents reside are usually not exactly identically distributed. The slight difference in the stochastic properties of the transition models for multiple agents could cause the learning convergence issue. One possible method to address this problem is by adjusting the frequency of global aggregation, i.e., after each global aggregation, a period of time is left for each agent to fine-tune its local parameters according to its own environment. Apart from the non-identical environment problem, another interesting and important problem is how to leverage FL to make RL algorithms converge better and faster. It is well-known that DRL algorithms could be unstable and diverge, especially when off-policy training is combined with function approximation and bootstrapping. In FRL, the training curves of some agents could diverge while others converge although the agents are trained in the exact replicas of the same environment. By leveraging FL, it is envisioned that we could expedite the training process as well as increase the stability. For example, we could selectively aggregate the parameters of a subset of agents with a larger potential for convergence, and later transfer the converged parameters to all the agents. To tackle the above problems, several possible solutions proposed for FL algorithm contains certain reference significance. For example, server operators could account for heterogeneity inherent in partial information by adding a proximal term \cite{DBLP:journals/corr/abs-1812-06127}. The local updates submitted by agents are constrained by the tunable term and have a different effect on the global parameters. In addition, a probabilistic agent selection scheme can be implemented to select the agents whose local FL models have significant effects on the global model to minimize the FL convergence time and the FL training loss \cite{9292468}. Another problem is theoretical analysis of the convergence bounds. Although some existing studies have been directed at this problem \cite{li2020convergence}, the convergence can be guaranteed since the loss function is convex. How to analyze and evaluate the non-convex loss functions in HFRL is also an important research topic in the future.

	\subsection{Agents without rewards in VFRL}
	In most existing works, all the RL agents have the ability to take part in full interaction with the environment and can generate their own actions and rewards. Even though some MARL agents may not participate in the policy decision, they still generate their own reward for evaluation. In some scenarios, special agents in VFRL take the role of providing assistance to other agents. They can only observe the environment and pass on the knowledge of their observation, so as to help other agents make more effective decisions. Therefore, such agents do not have their own actions and rewards. The traditional RL models cannot effectively deal with this thorny problem. Many algorithms either directly use the states of such agents as public knowledge in the system model or design corresponding action and reward for such agents, which may be only for convenience of calculation and have no practical significance. These approaches cannot fundamentally overcome the challenge, especially when privacy protection is also an essential objective to be complied with. Although the FedRL algorithm \cite{zhuo2020federated} is proposed to deal with the above problem, which has demonstrated good performance, there are still some limitations. First of all, the number of agents used in experiments and algorithms is limited to two, which means the scalability of this algorithm is not high and VFRL algorithms for a large number of agents need to be designed. Secondly, this algorithm uses Q-network as the federated model, which is a relatively simple algorithm. Therefore, how to design VFRL models based on other more complex and changeable networks remains an open issue.
	
	\subsection{Communications}
	In FRL, the agents need to exchange the model parameters, gradients, intermediate results, etc., between themselves or with a central server. Due to the limited communication resources and battery capacity, the communication cost is an important consideration when implementing these applications. With an increased number of participants, the coordinator has to bear more network workload within the client-server FRL model  \cite{DBLP:journals/corr/abs-1902-01046}. This is because each participant needs to upload and download model updates through the coordinator. Although the distributed peer-to-peer model does not require a central coordinator, each agent may have to exchange information with other participants more frequently. In current research for distributed models, there are no effective model exchange protocols to determine when to share experiences with which agents. In addition, DRL involves updating parameters in deep neural networks. Several popular DRL algorithms, such as DQN \cite{mnih_human-level_2015} and DDPG \cite{lillicrap2019continuous}, consist of multiple layers or multiple networks. Model updates contain millions of parameters, which isn't feasible for scenarios with limited communication resources. The research directions for the above issues can be divided into three categories. First, it is necessary to design a dynamic update mechanism for participants to optimize the number of model exchanges. A second research direction is to use model compression algorithms to reduce the amount of communication data. Finally, aggregation algorithms that allow participants to only submit the important parts of local model should be studied further.
	
	
	\subsection{Privacy and Security}
	Although FL provides privacy protection that allows the agents to exchange information in a secure manner during the learning process, it still has several privacy and security vulnerabilities associated with communication and attack \cite{DBLP:journals/corr/abs-2003-02133}. As FRL is implemented based on FL algorithms, these problems also exist in FRL in the same or variant form. It is important to note that the data poisoning attack is a different attack mode between FL and FRL. In the existing classification tasks of FL, each piece of training data in the dataset corresponds to a respective label. The attacker flips the labels on training examples in one category onto another while the features of the examples are kept unchanged, misguiding the establishment of a target model \cite{DBLP:journals/corr/abs-1808-04866}. However, in the decision-making task of FRL, the training data is continuously generated from the interaction between the agent and the environment. As a result, the data poisoning attack is implemented in another way. For example, the malicious agent tampers with the reward, which causes the evaluative function to shift. An option is to conduct regular safety assessments for all participants. Participants whose evaluation indicator falls below the threshold are punished to reduce the impact on the global model \cite{anwar2021multitask}. Apart form the insider attacks which are launched by the agents in the FRL system, there may be various outsider attacks which are launched by intruders or eavesdroppers. Intruders may hide in the environment where the agent is and manipulate the transitions of environment to achieve specific goals. In addition, by listening to the communication between the coordinator and the agent, the eavesdropper may infer sensitive information from exchanging parameters and gradients \cite{DBLP:journals/corr/abs-1906-08935}. Therefore, the development of technology that detects and protects against attacks and privacy threats does have great potential and is urgently needed.
	
	\subsection{Join and exit mechanisms design}
	One overlooked aspect of FRL-based research is the join and exit process of participants. In practice, the management of participants is essential to the normal progression of cooperation. As mentioned earlier in the security issue, the penetration of malicious participants severely impacts the performance of the cooperative model and the speed of training. The joining mechanism provides participants with the legal status to engage in federated cooperation. It is the first line of defense against malicious attackers. In contrast, the exit mechanism signifies the cancellation of the permission for cooperation. Participant-driven or enforced exit mechanisms are both possible. In particular, for synchronous algorithms, ignoring the exit mechanism can negatively impact learning efficiency. This is because the coordinator needs to wait for all participants to submit their information. In the event that any participant is offline or compromised and unable to upload, the time for one round of training will be increased indefinitely. To address the bottleneck, a few studies consider updating the global model using the selected models from a subset of participants \cite{8761315, wang_optimizing_2020}. Unfortunately, there is no comprehensive consideration of the exit mechanism, and the communication of participants is typically assumed to be reliable. Therefore, research gaps of FRL still exist in joining and exiting mechanisms. It is expected that the coordinator or monitoring system, upon discovering a failure, disconnection, or malicious participant, will use the exit mechanism to reduce its impact on the global model or even eliminate it.

	\subsection{Incentive mechanisms}
	For most studies, the agents taking part in the FRL process are assumed to be honest and voluntary. Each agent provides assistance for the establishment of the cooperation model following the rules and freely shares the masked experience through encrypted parameters or gradients.  An agent's motivation for participation may come from regulation or incentive mechanisms. The FRL process within an organization is usually governed by regulations. For example, BSs belonging to the same company establish a joint model for offloading and caching. Nevertheless, because participants may be members of different organizations or use disparate equipment, it is difficult for regulation to force all parties to share information learned from their own data in the same manner. If there are no regulatory measures, participants prone to selfish behavior will only benefit from the cooperation model but not submit local updates. Therefore, the cooperation of multiple parties, organizations, or individuals requires a fair and efficient incentive mechanism to encourage their active participation. In this way, agents providing more contributions can benefit more and selfish agents unwilling to share there learning experience will receive less benefit. As an example, Google Keyboard \cite{DBLP:journals/corr/abs-1812-02903} users can choose whether or not to allow Google to use their data, but if they do, they can benefit from more accurate word prediction. Although an incentive mechanism in a context-aware manner among data owners is proposed in the study from Yu \etal \cite{10.1145/3375627.3375840}, it is not suitable for the RL problems. There is still no clear plan of action regarding how the FRL-based application can be designed to create a reasonable incentive mechanism for inspiring agents to participate in collaborative learning. To be successful, future research needs to propose a quantitative standard for evaluating the contribution of agents in FRL.
	
	\subsection{Peer-to-peer cooperation}
	FRL applications have the option of choosing between a central server-client model as well as a distributed peer-to-peer model. A distributed model can not only eliminate the single point of failure, but it can also improve energy efficiency significantly by allowing models to be exchanged directly between two agents. In a typical application, two adjacent cars share experience learned from road condition environment in the form of models with D2D communications to assist autonomous driving. However, the distributed cooperation increases the complexity of the learning system and imposes stricter requirements for application scenarios. This research should include, but not be limited to, the agent selection method for the exchange model, the mechanism for triggering the model exchange, the improvement of algorithm adaptability, and the convergence analysis of the aggregation algorithm.

	
	
	
	\section{Conclusion}
	As a new and potential branch of RL, FL can make learning safer and more efficient while leveraging the benefits of FL. We have discussed the basic definitions of FL and RL as well as our thoughts on their integration in this paper. In general, FRL algorithms can be classified into two categories, \ie HFRL and VFRL. Thus, the definition and general framework of these two categories have been given. Specifically, we have highlighted the difference between HFRL and VFRL. Then, a lot of existing FRL schemes have been summarized and analyzed according to different applications. Finally, the potential challenges in the development of FRL algorithms have been explored. Several open issues of FRL have been identified, which will encourage more efforts toward further research in FRL.
	
	%
	%
	%
	%
	%
	%
	%
	%
	%

	

	
	
	
	
	\bibliographystyle{IEEEtran}
	\bibliography{reference}

\end{document}